\newcounter{myctr}
\def\myitem{\refstepcounter{myctr}\bibfont\parindent0pt\hangindent13pt\themyctr.\enskip}
\def\myhead#1{\vskip10pt\noindent{\bibfont #1:}\vskip4pt}
\documentclass{ws-ijhr}

\usepackage{graphicx} % for pdf, bitmapped graphics files
\usepackage{times} % assumes new font selection scheme installed
\usepackage{amsmath} % assumes amsmath package installed
\usepackage{amssymb}  % assumes amsmath package installed
\usepackage{multirow,color}
\usepackage{cite}
\usepackage{algorithm}
\usepackage{algpseudocode}
\usepackage{varwidth}
\usepackage{subfig}
\usepackage{microtype}

\usepackage[hyphens]{url}
\usepackage[font=small,labelfont=bf]{caption}
\usepackage{color}

\DeclareMathAlphabet{\bdmath}{OML}{cmm}{b}{it}     % Same as latex boldmath
\mathversion{normal}

\def\0{\mathbf{0}}

\sloppypar

\DeclareCaptionLabelFormat{andtable}{#1~#2 \& \tablename~\thetable}

\begin{document}

\catchline{}{}{}{}{}
%
%%%%%%%%%%%%%%%%%%%%%%%%%%%%%%%%%%%%%%%%%%%%%%%%%%%%%%%%%%%%%%%%%%%%
%\title{Data-driven Inference of Object Properties from a Tactile Sensing Array Given Varying Joint Stiffness and Velocity}
\title{Inferring Object Properties with a Tactile Sensing Array Given Varying Joint Stiffness and Velocity}

%INFERRING OBJECT PROPERTIES WITH A TACTILE SENSING ARRAY GIVEN VARYING JOINT STIFFNESS AND VELOCITY}

%\title{GENERALIZING INFERENCE OF OBJECT PROPERTIES BASED ON TACTILE SENSING ACROSS VARYING ROBOT STIFFNESSES AND VELOCITIES}
%\ck{at minimum, this needs to clarify that the robot's stiffness and velocity is being varied. it would also be nice if the title clarified that a robot is making contact with an object and trying to infer its properties based on tactile sensing.}

\author{Tapomayukh Bhattacharjee$^\ast$, James M. Rehg$^\dagger$, and Charles C. Kemp$^\ddagger$}

\address{Institute of Robotics and Intelligent Machines,\\
Georgia Institute of Technology,\\
Atlanta, GA-30332, USA\\
$^\ast$tapomayukh@gatech.edu\\
$^\dagger$rehg@cc.gatech.edu\\
$^\ddagger$charlie.kemp@bme.gatech.edu}

% \author{JAMES M. REHG}

% \address{Institute of Robotics and Intelligent Machines,\\
% Georgia Institute of Technology,\\
% Atlanta, GA-30332, USA\\
% rehg@cc.gatech.edu}

% \author{CHARLES C. KEMP}

% \address{Institute of Robotics and Intelligent Machines,\\
% Georgia Institute of Technology,\\
% Atlanta, GA-30332, USA\\
% charlie.kemp@bme.gatech.edu}

\maketitle

\begin{history}
\received{1st April 2017} %
%\revised{Day Month Year}  %
%\accepted{Day Month Year} %
%\comby{(xxxxxxxxxx)}
\end{history}

\begin{abstract}
%The abstract should summarize the context, content and conclusions of the paper in less than 200 words. It should not contain any references or displayed equations.

Whole-arm tactile sensing enables a robot to sense contact and infer contact properties across its entire arm. Within this paper, we demonstrate that using data-driven methods, a humanoid robot can infer mechanical properties of objects from contact with its forearm during a simple reaching motion. A key issue is the extent to which the performance of data-driven methods can generalize to robot actions that differ from those used during training. To investigate this, we developed an idealized physics-based lumped element model of a robot with a compliant joint making contact with an object. Using this physics-based model, we performed experiments with varied robot, object and environment parameters. We also collected data from a tactile-sensing forearm on a real robot as it made contact with various objects during a simple reaching motion with varied arm velocities and joint stiffnesses. The robot used one nearest neighbor classifiers (1-NN), hidden Markov models (HMMs), and long short-term memory (LSTM) networks to infer two object properties (hard vs. soft and moved vs. unmoved) based on features of time-varying tactile sensor data (maximum force, contact area, and contact motion). We found that, in contrast to 1-NN, the performance of LSTMs (with sufficient data availability) and multivariate HMMs successfully generalized to new robot motions with distinct velocities and joint stiffnesses. Compared to single features, using multiple features gave the best results for both experiments with physics-based models and a real-robot.
\end{abstract}


%%%%% now, begins the document %%%%%%%%%%%%%%%%%%%%%%%%%%%%%%%%%%%%%%%%%%%%%%
%Contributions to the {\it International Journal of Humanoid Robotics} should be about 15 to 20 printed pages long (not a requirement - confirmed by editor), but shorter communications and longer reviews will also be considered for publication. Authors are encouraged to have their contribution checked for grammar. American spelling should be used.
%%%%%%%%%%%%%%%%%%%%%%%%%%%%%%%%%%%%%%%%%%%%%%%%%%%%%%%%%%%%%%%%%%%%%%%%%%%%%
%%
%%  SECTION : Introduction
%%
%%%%%%%%%%%%%%%%%%%%%%%%%%%%%%%%%%%%%%%%%%%%%%%%%%%%%%%%%%%%%%%%%%%%%%%%%%%%%
%%%%%%%%%%%%%%%%%%%%%%%%%%%%%%%%%%%%%%%%%%%%%%%%%%%%%%%%%%%%%%%%%%%%%%%%%%%%%%%%
\section{Introduction}\label{sec:intro}
Manipulation in unstructured environments with high clutter is difficult due to a variety of factors, including uncertainty about the state of the world, a lack of non-contact trajectories, and reduced visibility for line-of-sight sensors~\cite{jain2013reaching}. Tactile sensing is well-matched to these challenges, since it benefits from contact and uses sensors that move with the manipulator into the clutter. When contact occurs with tactile sensors, the robot has an opportunity to acquire information. By fully covering the robot's manipulator with tactile sensors, the robot is likely to have more opportunities to acquire useful information through contact. However, with a typical serial manipulator, a robot cannot independently control the pose of each of the sensors. In addition, contact may not be anticipated.

Within this paper, we address the problem of haptic perception based on 
%\ck{it's risky to use the word "simple", so i deleted it. also few things about the real world are truly simple} 
contact~\cite{BhattacharjeeRehgKemp2012, BhattacharjeeKapustaRehgKemp2013, BhattacharjeeGriceKempSystems2014} with a tactile-sensing forearm on a humanoid robot. The time varying signals from a tactile sensing array depend on the robot's actions, including the joint stiffnesses and joint velocities of the robot. A key problem for data-driven approaches is how to infer object properties based on these signals when the robot's actions are different from those used during training. In other words, after a robot has learned about an object using one action, it would ideally be able to infer the same properties of the object when making contact with it using a different action. In this paper, we consider an example of this type of problem. \textit{Specifically, we focus on a robot inferring object properties with a tactile sensing array when the robot's joint stiffness and joint velocity differ from those used during training.}

\begin{figure}[t!]
\centering
\includegraphics[height=5cm]{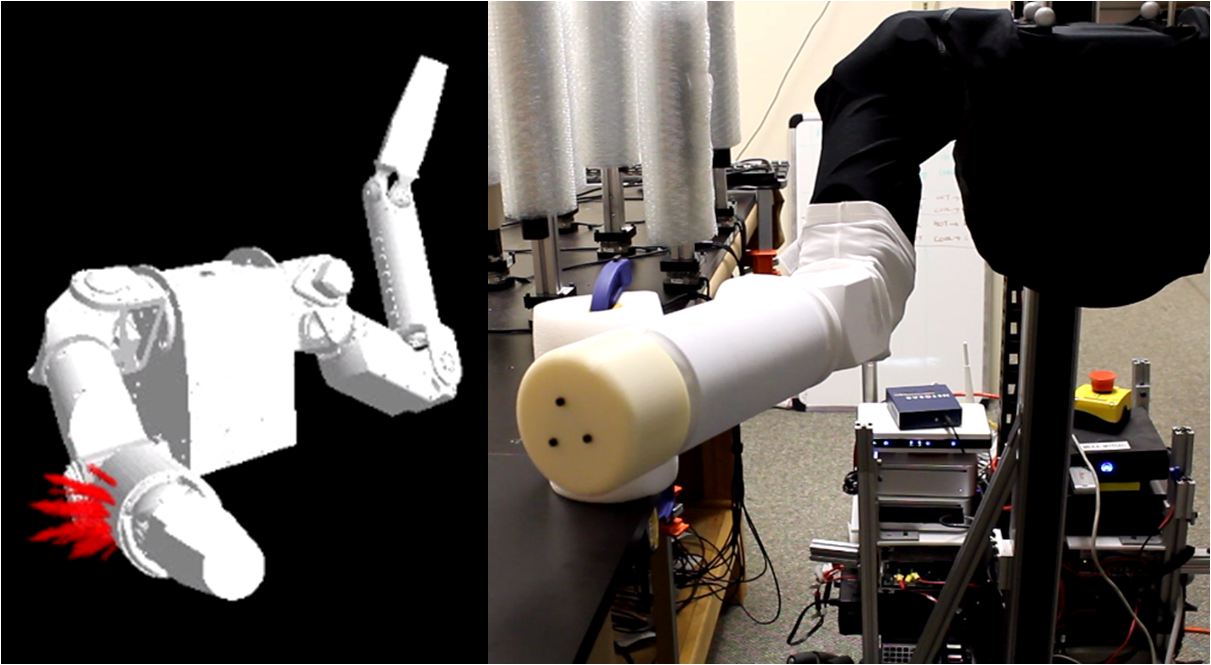}
\caption{\label{fig:cody}\textit{Force data from forearm skin sensor mounted on the robot, Cody. The forearm came into contact with a cylindrical object made of polystyrene foam, while trying to reach a goal configuration. The red arrows show the forces acting on the skin.}}
\end{figure}

The type of action we consider in this work is a short compliant movement of a robot's forearm akin to movements that occur during reaching.  Figure \ref{fig:cody} shows one such example when the robot, Cody, came into contact with a cylindrical object made of polystyrene foam, while trying to reach a goal configuration. We intentionally do not have the robot use exploratory behaviors, and instead investigate inference from short duration contact (i.e., 5\,s and 1.2\,s). Our choice of action is inspired by the potential for robots to infer useful properties of the world based on \textit{incidental contact}. By incidental contact, we mean contact that is not central to the robot's current actions and may occur unexpectedly or unintentionally.\footnote{This description supersedes our previous descriptions from~\cite{BhattacharjeeKapustaRehgKemp2013} and~\cite{BhattacharjeeGriceKempSystems2014}.} For example, while manipulating in cluttered environments, incidental contact is more likely to occur and can be common with some approaches to robot control ~\cite{jain2013reaching,whc_2013,icorr_2013}.  Incidental contact will typically not involve active exploration and interrogation of the contact, since the robot will be directing its resources elsewhere~\cite{BhattacharjeeKapustaRehgKemp2013, BhattacharjeeGriceKempSystems2014}. Each such contact event is an opportunity for sensing. Unlike deliberate probing, during which the robot has more control over its actions to optimize its sensing, the sensing is opportunistic during these motions and contact events can be brief and simple, which could make inference challenging.

\subsection{Opportunities and Challenges of Haptic Perception}\label{ssec:obj_properties}
Inferring mechanical properties of objects from contact could be beneficial in a number of ways. For example, we have shown that haptically recognizing leaves vs. trunk while a robot reaches into artificial foliage can be used to haptically map the environment and plan paths to goals \cite{BhattacharjeeKapustaRehgKemp2013, BhattacharjeeGriceKempSystems2014}. Rather than recognizing a particular object type, detecting an object's properties could be informative in novel environments. Detecting that an object has \textit{moved} due to a robot's actions might be used by the robot to make better decisions, such as moving the object away in order to access a new location, or avoiding the object, so as not to alter the environment further. Likewise, detecting that an object is \textit{hard} or \textit{soft} has implications for the robot's ability to compress the object and the consequences of collisions with the object.

One of the challenges of haptics is that sensing depends on action. For example, haptic perception of surface roughness depends on contact speed \cite{lederman1999perceiving} and contact force \cite{lederman2000perceiving}. Many existing tactile systems use carefully controlled exploratory behaviors to reduce variability of the actions and gather information about the object (See Section \ref{sec:lit_review}). The signals produced from a robot's tactile sensors depend on the mechanics of both the object and the robot. Physical interaction depends on the dynamics of impact between a robot and an object. For example, \textit{depending upon how stiff or compliant a robot is or how fast a robot moves, the physical interaction will vary.} As such, a key issue for data-driven approaches to tactile perception of contact is the extent to which the performance of perceptual classifiers can generalize when a robot's actions differ from those used to collect training data.

%\ck{Probably here, but possibly elsewhere, you should provide more insight into why hard/soft, moved/unmoved with varying robot joint stiffness and velocity is challenging. You should give more concrete examples about the ambiguities that arise and note the confusions that showed up in our results that reinforce the challenge. You've conveyed that this problem could be useful, but you need to do a better job establishing that it is challenging and intellectually interesting.}

Classifying an object based on its compliance or mobility can become challenging when robot joint stiffness or velocity changes. For example, interaction forces are a function of the robot joint stiffness as well as object stiffness. The forces generated by contact between a stiff robot and a hard object is not the same as the forces generated by contact between a compliant robot and a hard object. Hence, perceiving an object as hard/soft is challenging if the robot stiffness changes, because it is difficult to distinguish the interaction between a stiff robot and a soft object and a compliant robot and a hard object. Similarly, it may be difficult to determine if an object has moved based on tactile sensing if the robot actions change. A movement in the robot's arm after contact can be due to the actual movement of the object (e.g. sliding motion) or because of the compliance in the structure of the object (the robot arm pushes into the object and the object deforms). A compliant robot moving with a low velocity could push into a soft object resulting in robot arm movement after initial contact, which could be similar to the actual movement of a hard object (without any deformation) generated by a stiff robot impacting with a high velocity. In this work, we focus on addressing these challenges by:

\begin{itemize}
\item{classifying an object into one of the four categories: 1)~\textit{Hard-Unmoved}~(HU), 2)~\textit{Hard-Moved}~(HM), 3)~\textit{Soft-Unmoved}~(SU), and 4)~\textit{Soft-Moved}~(SM), and}
\item{investigating the potential of data-driven methods to generalize the performance of haptic perception to different robot conditions such as robot stiffness and velocity used to collect training data.}
\end{itemize}

\subsection{Methods}\label{ssec:methods}
We used univariate and multivariate HMMs as well as long short-term memory (LSTM) networks to infer object properties (Section \ref{sec:approach}) and compared the results with 1-NN used in our previous work \cite{BhattacharjeeRehgKemp2012}. In Section \ref{sec:simulations}, we present an idealized physics-based lumped element model of a robot with a compliant joint making contact with an object. Using this model, we performed experiments with physics-based simulations of varied robot, object, and environment parameters in Section \ref{sec:sims}. In Section \ref{sec:exps}, we present experiments with a real robot for which we varied the robot arm's velocity and joint stiffness to values distinct from those used during training. Our multivariate HMM-based method performed well in experiments with both physics-based simulations (Section \ref{sec:results_sim}) and with a real robot (Section \ref{sec:results}) for classifying objects into the four categories. Our LSTM networks performed better when a larger amount of data was available with the physics-based simulation compared to experiments with the real robot. Section \ref{ssec:limitations} discusses the methods and their limitations and Section \ref{sec:conclusion} provides the conclusion of our work.

% \subsection{Contributions}\label{ssec:contributions}
% \ck{if you're going to explicitly state the contributions, which you don't have to. then, i think you need to do a better job of it. }
% In summary, our main contributions are:
% \begin{itemize}
% \item{We developed methods to infer properties of objects based on whole-arm tactile sensing during a simple reaching motion.}
% \item{We developed a physics-based lumped element model for modeling robot-object physical interactions. This model enabled us to generate simulated data for a wide variety of robot, object, and environment conditions, which would otherwise be challenging to collect using experiments with a real robot. \ck{this contribution doesn't sufficiently note the success of the model, which resulted in similar results as the real robot}}
% \item{We developed methods to test under varying robot motions using both the physics-based models and a real robot. We also analyzed the role of different state-of-the-art algorithms and tactile features for classification performance. Using our methods, the classification performance generalized well to different robot stiffnesses and velocities. \ck{i don't really understand this claimed contribution}}
% \end{itemize}

%%%%%%%%%%%%%%%%%%%%%%%%%%%%%%%%%%%%%%%%%%%%%%%%%%%%%%%%%%%%%%%%%%%%%%%%%%%%%
%%
%%  SECTION : Related Work
%%
%%%%%%%%%%%%%%%%%%%%%%%%%%%%%%%%%%%%%%%%%%%%%%%%%%%%%%%%%%%%%%%%%%%%%%%%%%%%%
\section{Related Work}\label{sec:lit_review}
Object categorization is a well studied task. In this work, we focus on inferring properties of objects based on haptic sensing. Although there have been multiple studies on haptics-based compliance discrimination, most have used specific exploratory behaviors using end effectors to extract information from the environment. Lederman and Klatzky discussed in detail the various factors and exploratory behaviors that humans use for inferring haptic properties of objects~\cite{lederman1993extracting}. However, studies of discrimination tasks using information from incidental contact with large-area tactile sensing are lacking. Also, studies on inferring properties of objects and \textit{generalizing the performance across various conditions such as robot stiffnesses and velocities} are rare.

Researchers have also used haptics and tactile sensing to infer properties of the world for purposes other than categorizing objects. Silvera-Tawil et al. presented a comprehensive review of state-of-the-art methods in tactile sensing for robots in socially interactive scenarios~\cite{silvera2015artificial}. Wu et al. used tactile arrays for recognizing human intended directions in two dimensions using support vector machine classifiers (SVMs)~\cite{wu2013two}. Muscari et al. developed algorithms for reconstructing force and shape distributions using capacitive tactile sensing arrays~\cite{muscari2013real}. Hughes et al. presented a soft robotic artificial skin for texture recognition and localization~\cite{hughes2015texture}. Javaid et al. used pressure sensors to classify human activities during assistive tasks to help older adults~\cite{javaid2015using}. Matheus and Dollar used a load cell and a custom-built device to infer static friction properties of some `Objects of Daily Living'~\cite{matheus2010benchmarking}. Boonvisut and {\c{C}}avu{\c{s}}oglu used exploratory behaviors and vision to collect deformation data for identifying boundary constraints of deformable objects required to estimate soft tissue properties~\cite{boonvisut2014identification}. Researchers have used tactile sensors to classify events such as slips between fingertips and objects and slips between objects and external surfaces~\cite{heyneman2015slip, schurmann2012high, ho2014slip}. Researchers have also used haptic sensing for texture perception \cite{kaboli2015hand, SukhoySahaiSinapovStoytchev2009, howe1993dynamic}, tactile servoing \cite{li2013control}, contour following \cite{martinez2013active}, and human-robot collaborative tasks \cite{agravante2014collaborative, calinon2009learning}.

In the following subsections, we review the existing literature that addresses object categorization using haptics.

\subsection{Material Property Based Classification}\label{subsec:material}
In this work, we do not explicitly model material properties, but the features that we extract from the interactions between the robot arm and environmental objects are a direct consequence of material properties. Drimus et al. classified hard and deformable objects based on haptic feedback from a novel tactile sensor consisting of a flexible, piezo-resistive rubber~\cite{DrimusKootstraBilbergKragic2011}. They represented tactile information from a palpation procedure as a time series of features and used a k-nearest-neighbor (k-NN) classifier to categorize the objects~\cite{DrimusKootstraBilbergKragic2011}. Our classification scheme considers both compliance and mobility characteristics and uses information from incidental contact sensed with large-area tactile sensors. 

Chu et al. presented research that uses discrete HMMs to construct a feature vector of likelihoods and used binary SVM classifiers to classify those vectors and automatically assign 24 adjectives to 60 objects~\cite{chu2015robotic}. In contrast to our work, their research focused on classifying data using both static and dynamic features from four deliberate exploratory procedures with sophisticated BioTac \cite{lin2009signal} robotic fingers from~\textit{Syntouch}. 

Kaboli et al. used multi-modal tactile features to distinguish texture and weight of objects using sliding and non-sliding exploratory behaviors \cite{kaboli2014humanoids}. Kim and Kesavadas presented a method for estimating the material properties of objects (steel, aluminum, wood, silicon rubber) using an active tapping procedure~\cite{KimKesavadas2006}. Takamuku et al. estimated the hardness properties of objects through tapping and squeezing behaviors~\cite{TakamukuGomezHosodaPfeifer2007}. Hosoda and Iwase obtained haptic data using a bionic hand to grip an object. They used a recurrent neural network to classify objects based on haptic cues learned from dynamic interactions~\cite{HosodaIwase2010}. Nizar et al. classified the material type and surface properties by developing a sensor that used a lightweight plunger probe to detect surface properties. They also used an optical mouse sensor to obtain surface images and implemented a radial basis function neural network for classification~\cite{NizarCharniyaDudul2008}. Liarokapis et al. used random forests to classify size and stiffness of objects as well as distinguish object types using a single force closure grasp with an underactuated robotic hand and force sensors~\cite{liarokapis2015unplanned}. Schmitz et al. used power grasping of objects and multiple modalities for object recognition with deep learning~\cite{schmitz2014tactile}. Kiwatthana and Kaitwanidvilai used system identification and K-means clustering techniques to classify different cans using proprioceptive feedback~\cite{kiwatthana2014development}. Hoelscher et al. used BioTac fingers and Schunk F/T sensors with multiple classifiers and feature extraction methods for object recognition. They concluded that simple, dimensionally-reduced features performed better than more elaborate features~\cite{hoelscher2015evaluation}. 

In summary, although there have been many studies on material property based classification, most have focused on carefully controlled specific exploratory behaviors using the robot's end-effector. These studies have not looked at whether the classification performance can generalize to different robot behaviors.

\subsection{Shape Based Classification}\label{subsec:shape}
Many researchers have either used tactile images from touch sensors or analyzed object shape deformation behaviors for object categorization. Schneider et al. applied a ``bag-of-words" approach and unsupervised clustering techniques to categorize objects~\cite{SchneiderSturmStachniss2009}. Pezzementi et al. identified the principal components of features, then clustered them, and constructed per-class histograms as a class characteristic~\cite{PezzementiPlakuReydaHager2011}. Gorges et al. introduced passive joints in the hand for better adaptibility to different object shapes and used a Bayes classifier to classify the objects~\cite{GorgesNavarroGogerWorn2010}. Babu et al. used `C4.5' algorithm to generate a decision tree and a naive Bayes classifier to categorize shapes of objects using a tactile sensor array~\cite{babu2014machine}.

Other researchers have analyzed deformation behavior to classify objects. They have used vision and haptic sensors~\cite{UedaHiraiTanaka2004, bjorkman2013enhancing} or finite element models~\cite{FrankSchmeddingStachnissTeschnerBurgard2010a, FrankSchmeddingStachnissTeschnerBurgard2010b} and volumetric models such as superquadrics~\cite{AllenRoberts1989}, polyhedral models~\cite{CaselliMagnaniniZanichelli1994} or wrapping polyhedra~\cite{FaldellaFringuelliPasseriRosi1997}.

To summarize, shape based classification methods have used tactile images or deformation behaviors to classify objects after exploring or grasping them using multi-fingered robot hands. Again, most of these studies used exploratory behaviors and have not looked at whether the performance can generalize to different robot behaviors used to collect training data.

\subsection{Functional Property Based Classification}\label{subsec:function}
This group of studies focused on functional property based classification or classification based on how objects behave when they are moved. Sinapov et al. used the acoustic properties of objects during specific interaction schemes and the behavioral interactions performed with them, such as grasping, shaking, dropping, pushing, and tapping, to classify 36 different household objects~\cite{SinapovWeimerStoytchev2009}. Berquist et al. monitored the changes in the joint torques of a robot while it performed five exploratory procedures --- lifting, shaking, crushing, dropping, and pushing --- on several objects and demonstrated that the robot could learn to recognize objects based solely on the joint-torque information~\cite{BerquistSchenckOhiriSinapovGriffithStoytchev2009}. Jain et al. used data-driven object centric models to haptically recognize specific doors as well as classes of doors (refrigerator vs. kitchen cabinet)~\cite{jain2013improving}. Griffith et al. used multiple exploratory behaviors and employed clustering techniques to categorize containers and non-containers. After extracting visual and acoustic features from interactions with objects, they employed unsupervised clustering techniques to form several categories~\cite{GriffithSinapovSukhoyStoytchev2011}. Sinapov et al. combined proprioceptive and auditory feedback and used a behavior-grounded relational classification model to recognize categories of household objects~\cite{SinapovBergquistSchenckOhiriGriffithStoytchev2011}. Sinapov and Stoytchev extended their previous work by using auditory, proprioceptive, and visual modalities to cluster novel and unlabeled objects for object individuation based on the robot's sensorimotor experience of handling those novel objects~\cite{sinapov2013grounded}.

To summarize, functional property based classification methods have used multimodal feedback from robot behaviors and actions to distinguish between different object categories. But, similar to other work, most of them used exploratory behaviors and have not looked at if the results can generalize to different robot behaviors.

\begin{figure}[t!]
\centering
\includegraphics[height=4.6cm]{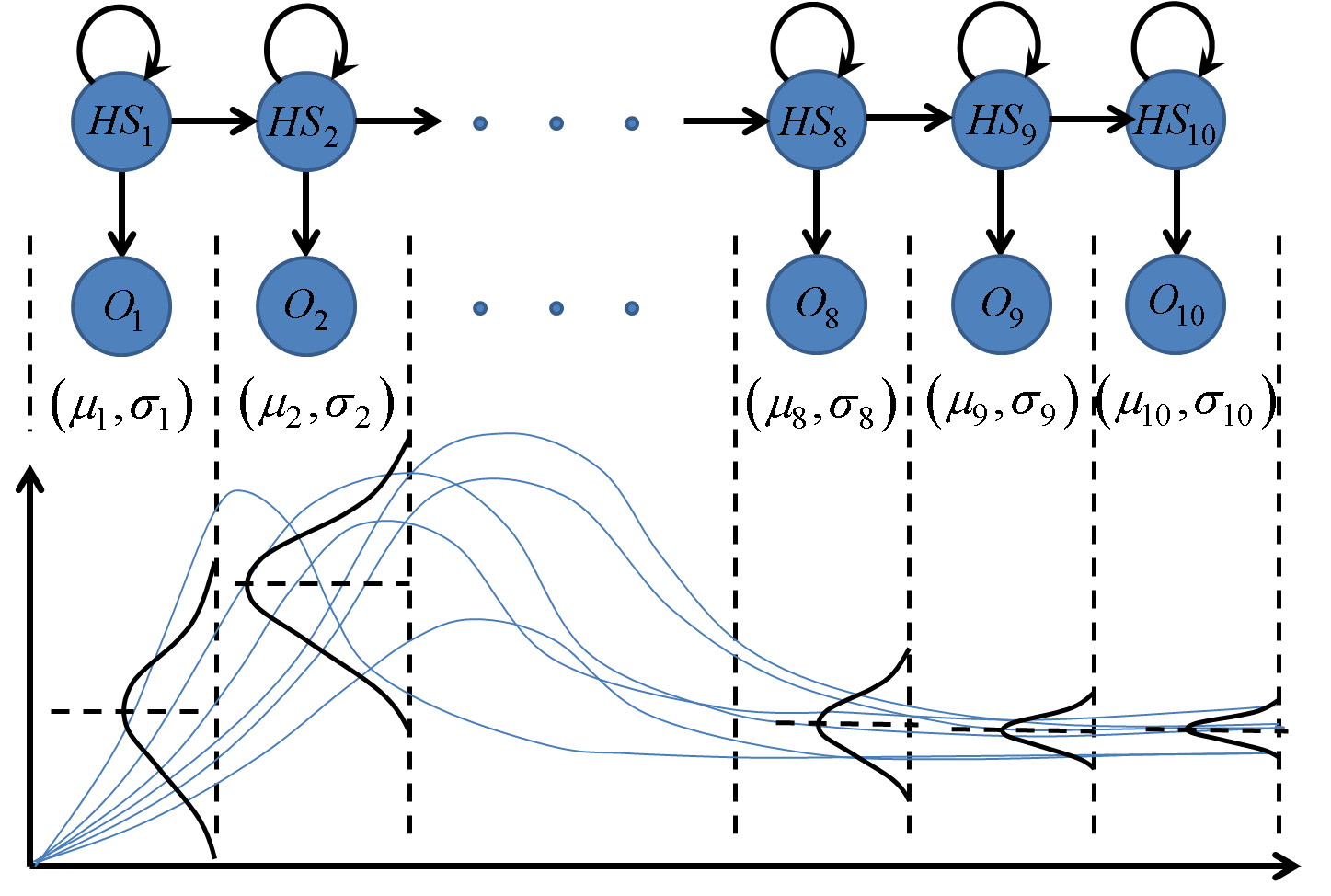}
\includegraphics[height=5.4cm]{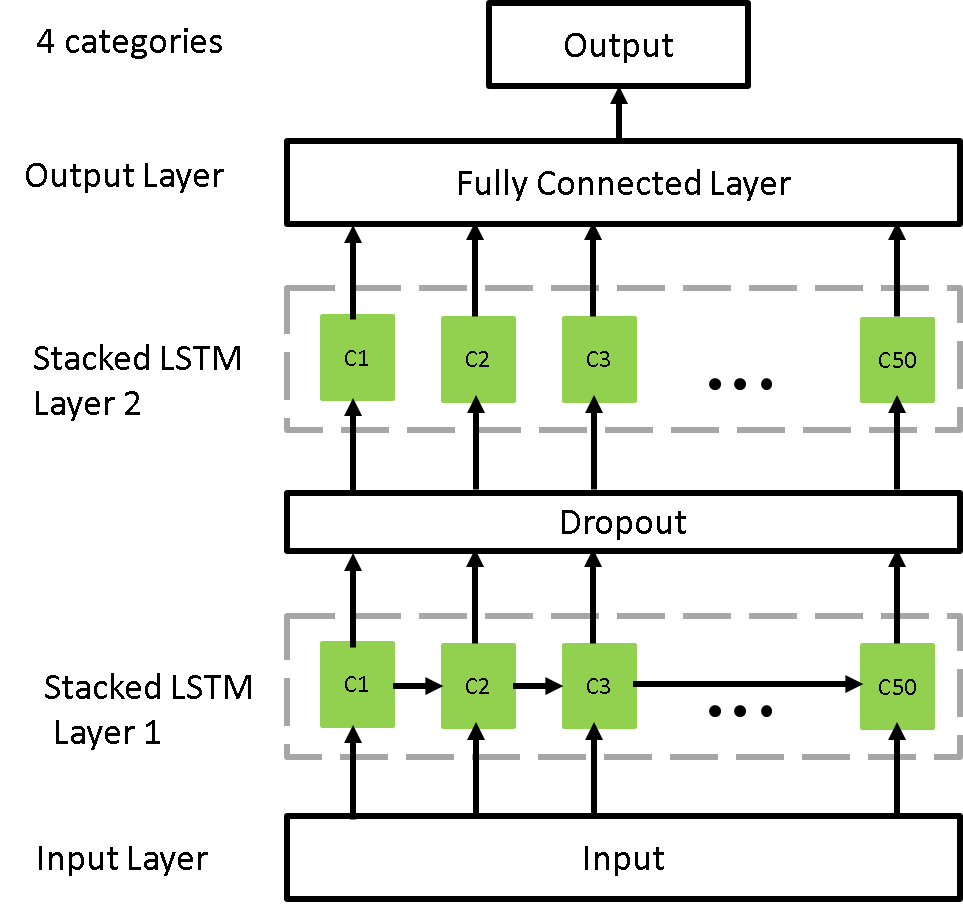}
\caption{\label{fig:lstms}\textit{Schematic of a left-right HMM with 10 states. The observations are modeled using Gaussian distributions (Left). Schematic of a Stacked LSTM network with 50 cells in each layer, dropout in between, and a fully connected layer as the output layer (Right).}}
\end{figure}

%%%%%%%%%%%%%%%%%%%%%%%%%%%%%%%%%%%%%%%%%%%%%%%%%%%%%%%%%%%%%%%%%%%%%%%%%%%%%
%%
%%  SECTION : Approach
%%
%%%%%%%%%%%%%%%%%%%%%%%%%%%%%%%%%%%%%%%%%%%%%%%%%%%%%%%%%%%%%%%%%%%%%%%%%%%%%
\section{Data-Driven Methods}\label{sec:approach}
For this work, we chose hidden Markov models and long short-term memory (LSTM) networks as our data-driven methods to infer object properties from contact. To compare our results, we used k-nearest neighbors from our previous work \cite{BhattacharjeeRehgKemp2012} as the baseline data-driven method.

\subsection{Hidden Markov Models}\label{ssec:approach_hmm}
Hidden Markov models (HMMs) have a long history of success for classifying time series such as human speech~\cite{Rabiner1990}. HMMs are known to have rich mathematical structure for modeling non-stationary signals and can work effectively in practical applications \cite{Rabiner1990}. In this study, we used HMMs to classify an object as being in one of four categories: 1)~\textit{Hard-Unmoved}~(HU), 2)~\textit{Hard-Moved}~(HM), 3)~\textit{Soft-Unmoved}~(SU), and 4)~\textit{Soft-Moved}~(SM).

A hidden Markov model (HMM) is a state-based data modeling tool that assumes the states are hidden and the system is a Markov process. The hidden states ($HS_i$) are inferred using observations ($O_t$) at time $t$. The components of an HMM include $N$, the number of states in the model; $\mathbf{A}$, the state transition probabilities; $\mathbf{B}$, the observation probabilities; and $\mathbf{\pi}$, the initial state probabilities. We use notation from \cite{Rabiner1990}. Equation (\ref{eq0}) shows the HMM model ($\lambda$). Eq. (\ref{eq1}) shows the state transition probabilities.

\begin{align}\label{eq0}
\begin{split}
\lambda  = \left( {\mathbf{A},\mathbf{B},\mathbf{\pi}} \right)\\
\end{split}
\end{align}

\begin{align}\label{eq1}
\begin{split}
\mathbf{A} = \left\{ {{a_{ij}}} \right\} = \left\{P\left( {{x_t} = HS_j|{x_{t - 1}} = HS_i}   \right)\right\}, ~1 \leq i,~j \leq N,
\end{split}
\end{align}

where $i$, $j$ are state indexes and

\begin{align}\label{eq1a}
\begin{split}
a_{ij} \geq 0, \\
\sum_{j=1}^{N}a_{ij} = 1
\end{split}
\end{align}

Note, for a left-right HMM, as time increases, the state index increases or stays the same. Therefore, there is an additional constraint given by

\begin{align}\label{eq1b}
\begin{split}
a_{ij} = 0,~j < i \\
\end{split}
\end{align}

Eq. (\ref{eq2}) represents the initial state probabilities.

\begin{align}\label{eq2}
\begin{split}
\mathbf{\pi}  = \left\{ {{\pi _i}} \right\} = \left\{P\left( {{x_1} = HS_i} \right)\right\}, ~1 \leq i \leq N
\end{split}
\end{align}

For a discrete HMM with $M$ distinct observation symbols, eq. (\ref{eq3}) shows the observation probabilities.

\begin{align}\label{eq3}
\begin{split}
\mathbf{B} = \left\{ {{b_j}\left( k \right)} \right\} = \left\{P\left( {O_t = v_k|{x_t} = HS_j} \right)\right\}, ~1 \leq j \leq N, \\
1 \leq k \leq M,
\end{split}
\end{align}

where $\mathbf{V} = \{v_1, v_2, ... , v_M\}$ are the individual symbols. However, for a continuous HMM (which we use in this work) with observation vector $\mathbf{O}$ and multivariate Gaussian emissions with mean vector $\boldsymbol{\mu}$ and covariance matrix $\mathbf{U}$, eq. (\ref{eq4}) shows the emission probabilities.

\begin{align}\label{eq4}
\begin{split}
\mathbf{B} = \left\{ {{b_j}\left( \mathbf{O} \right)} \right\} = \left\{ {P\left( {\mathbf{O} |{x_t} = HS_j} \right)} \right\} \\
= \left\{ \mathcal{N}\left[ \mathbf{O}, \boldsymbol{\mu} _j, \mathbf{U} _j \right] \right\},\\
1 \leq j \leq N
\end{split}
\end{align}

Figure \ref{fig:lstms} (Left) shows the schematic of one left-right HMM with $N=10$ states.

\subsection{Long Short-Term Memory Networks}\label{ssec:approach_lstm}
Long short-term memory (LSTM) networks have been successfully used for modeling time-series in many applications such as machine translation, generating cursive writing, and speech recognition \cite{sutskever2014sequence, graves2013generating, graves2013speech}. For mathematical details, please refer to \cite{hochreiter1997long}, where this was first introduced. LSTMs have also been successfully used for haptic perception such as, to estimate forces during robot-assisted dressing simulations \cite{zackory2017lstms}, and during robot-assisted surgery \cite{aviles2016exploring}. Note, Gao et al. \cite{gao2016deep} used LSTMs for haptic perception using visual and tactile features but the performance was lower compared to other deep-learning methods.

For our applications, we use an LSTM structure inspired by Gers et al. \cite{gers2000learning} in which each memory cell has an input gate, a forget gate, and an output gate. The specific structure that we used for our haptic classification tasks consists of 2 layers with 50 cells each. To reduce 
%\ck{reduce}
overfitting, we also added a dropout layer in between the two layers, which helps in regularization \cite{pham2014dropout}. We added a fully connected output layer. Section \ref{ssec:sim_preprocess} shows the details of the implementation. The LSTM has a total of 31,004 parameters. The 31,004 parameters correspond to the weights of the stacked LSTM layers, the individual cells, as well as the fully connected layer for the 4-category classification output. 

Figure \ref{fig:lstms} (Right) shows the schematic of the LSTM structure. 

%%%%%%%%%%%%%%%%%%%%%%%%%%%%%%%%%%%%%%%%%%%%%%%%%%%%%%%%%%%%%%%%%%%%%%%%%%%%%
%%
%%  SECTION : System Modeling
%%
%%%%%%%%%%%%%%%%%%%%%%%%%%%%%%%%%%%%%%%%%%%%%%%%%%%%%%%%%%%%%%%%%%%%%%%%%%%%%
\section{Physics-Based Model}\label{sec:simulations}
We modeled contact between a tactile-sensing robot forearm and an object using a lumped element model. We developed a physics-based model that can model the mechanics of a variety of robot-object physical interaction phenomena. In this work, the tactile-sensing robot arm moves by actuating its shoulder joint towards a goal angle. The arc formed by the contact point during the motion is approximately a straight line for small angle movements and a large radius. We modeled the robot arm's motion at the contact point as a linear movement. We modeled the object as a deformable object and the contact surface as flat. 

\subsection{Robot and Object Model}\label{ssec:bot_obj_model}
\begin{figure}[t!]
\centering
\includegraphics[height=3.4cm]{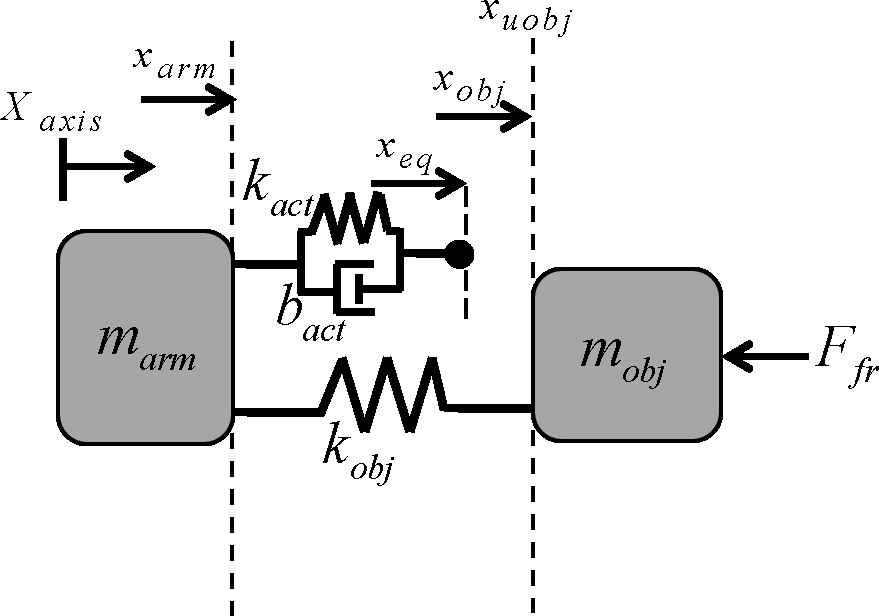}
\hfill
\includegraphics[height=3.4cm]{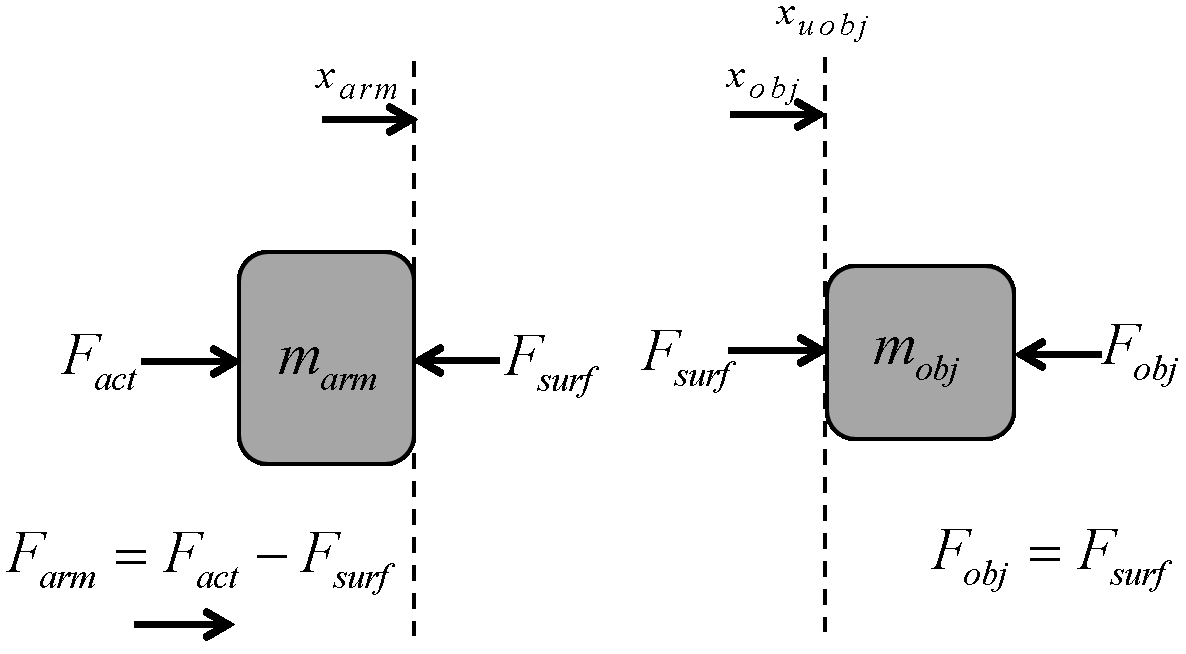}
\caption{\label{fig:sys_model}\textit{Lumped element model of our system at the onset of contact between the robot-arm and the object (Left). Free-body diagrams of the robot-arm and object in contact (Right).}}
\end{figure}

We modeled the robot's arm trajectory using equilibrium point control \cite{shadmehr1998equilibrium}. Figure \ref{fig:sys_model} (Left) shows a robot-arm of mass $m_{arm}$ making contact with an object of mass $m_{obj}$. $x _{arm}$ is the position of the robot-arm, $x _{obj}$ is the position of the object, and $x _{uobj}$ is the position of the undeformed object. During contact, if $\delta _{obj}$ is the object deformation, note that

\begin{equation}
{x _{obj}} = {x _{arm}} = {x _{uobj}} + \delta _{obj}, \label{eq:01a}
\end{equation}

otherwise, if the arm is not in contact with the object,

\begin{equation}
{x _{obj}} = {x _{uobj}} > {x _{arm}}, \label{eq:01b}
\end{equation}

$x _{eq}$ is the equilibrium point of the actuator spring with stiffness $k_{act}$ and actuator damping $b_{act}$. $k_{obj}$ is the object stiffness. $F_{fr}$ is the frictional force between the object and the environment.

Figure \ref{fig:sys_model} (Right) shows free-body diagrams for the system depicted in Fig. \ref{fig:sys_model} (Left). $F_{act}$ is the force applied on the robot-arm by the actuator, $F_{surf}$ is the force applied by the robot-arm to the surface of the object when in contact. $F_{arm}$ and $F_{obj}$ are the net forces acting on the robot-arm and the object respectively.

The force applied by the actuator to the robot-arm, $F_{act}$ is given by eq. (\ref{eq:1}).

\begin{equation}
{F_{act}} = {k_{act}}\left( {{x_{eq}} - {x_{arm}}} \right) + {b_{act}}\left( {\dot{x}_{eq} - \dot{x}_{arm}} \right), \label{eq:1}
\end{equation}

The net force on the arm, $F_{arm}$, is therefore calculated as in eq. (\ref{eq:2}). 

\begin{equation}
{F_{arm}} = {F_{act}} - {F_{surf}}, \label{eq:2}
\end{equation}

The net force on the object, $F_{obj}$, is given by eq. (\ref{eq:3}). 

\begin{equation}
{F_{obj}} = {F_{surf}} - {F_{fr}}, \label{eq:3}
\end{equation}

We model the joint as frictionless. Hence, the position of the robot-arm can be calculated by eq. (\ref{eq:4}), where ${\dot{x}_{{0_{arm}}}}$ is the initial velocity and ${x_{{0_{arm}}}}$ is the initial position of the arm. 

\begin{equation}
\begin{array}{l}
{\dot{x}_{arm}} = \int{{\left( {{F_{arm}}/{m_{arm}}} \right)dt} + {\dot{x}_{{0_{arm}}}}}, \\
{x_{arm}} = \int{\dot{x}_{arm}dt} + {x_{{0_{arm}}}}, \\
\end{array} \label{eq:4}
\end{equation}

The position of the undeformed object is calculated using eq. (\ref{eq:5}), where ${\dot{x}_{{0_{obj}}}}$ is the initial velocity and ${x_{{0_{obj}}}}$ is the initial position of the undeformed object.

\begin{equation}
\begin{array}{l}
{\dot{x}_{uobj}} = \int{{\left( {{F_{obj}}/{m_{obj}}} \right)dt} + {\dot{x}_{{0_{obj}}}}}, \\
{x_{uobj}} = \int{\dot{x}_{uobj}dt} + {x_{{0_{obj}}}}, \\
\end{array} \label{eq:5}
\end{equation}

$F_{surf}$ is calculated using eq. (\ref{eq:6}),

\begin{equation}
{F_{surf}} = {k_{obj}}\left( {x_{arm}} - {{x_{uobj}}} \right), \label{eq:6}
\end{equation}

Please note that the frictional force $F_{fr}$ is calculated differently depending on whether the object is moving or not as shown in eq. (\ref{eq:7}). If the applied force overcomes static friction, the object starts moving. ${\mu _s}$ and ${\mu _k}$ are the coefficients of static and kinetic friction, respectively and $g = 9.81$ m/s\textsuperscript{2}.

\begin{equation}
\begin{array}{l}
{F_{fr}} = \left\{ {\begin{array}{*{20}{c}}
   {{F_{surf}},\;stationary}  \\
   {{\mu _s}\left( {{m_{obj}}g} \right),\;just\;before\;motion}  \\
   {{\mu _k}\left( {{m_{obj}}g} \right),\;in\;motion}  \\
\end{array}} \right. \label{eq:7}
\end{array}
\end{equation}

\begin{figure}[t!]
\centering
\includegraphics[height=3.4cm]{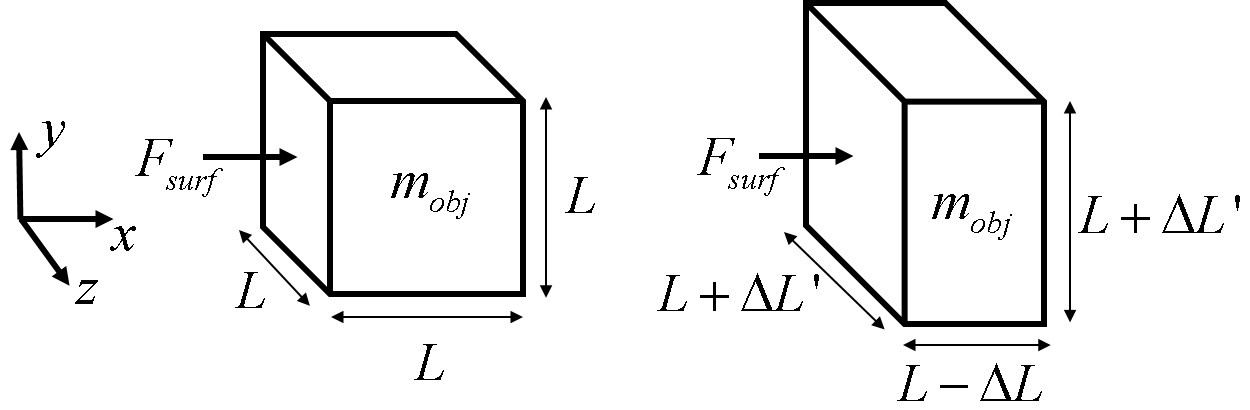}
\caption{\label{fig:area_model}\textit{Due to applied force in x-direction, an object undergoes axial compression. This results in elongation in the other orthogonal axes directions, thus increasing the surface contact area proportional to the object material's Poisson's ratio.}}
\end{figure}

\subsection{Contact Area Model}\label{ssec:area_model}
For an object in the shape of a cube with edge length $L_{obj}$, and surface area $A_{obj}$, let us assume that due to a robot-arm's applied force $F_{surf}$, there is a positive axial compression given by $\Delta L_{obj}$ in the x-direction (See Fig. \ref{fig:area_model}). Let the cube be made up of a homogeneous material with Poisson's ratio $\nu_{obj}$ \cite{westergaard1952theory} and due to the applied force, let the elongations in the y and z-directions be $\Delta L'_{obj}$. Before deformation, the surface area is given by,

\begin{equation}
A = {L^2},\label{eq:a1}
\end{equation}

After deformation, the surface area becomes

\begin{equation}
A' = {(L + \Delta L')^2},\label{eq:a2}
\end{equation}

Therefore, the increase in the surface area, due to the applied force, becomes

\begin{equation}
\Delta A = A' - A = {(L + \Delta L')^2} - {L^2},\label{eq:a3}
\end{equation}

which leads to

\begin{equation}
\frac{{\Delta A}}
{A} = {\left( {1 + \frac{{\Delta L'}}
{L}} \right)^2} - 1.\label{eq:a4}
\end{equation}

Let the strains in x, y, and z-axis due to the applied force $F_{surf}$ be given by $d\epsilon _x = \frac{dx}{x}$, $d\epsilon _y = \frac{dy}{y}$, and $d\epsilon _z = \frac{dz}{z}$ respectively. We assume that when the object is compressed, the transverse strain is positive.

\begin{equation}
\nu_{obj}  =  - \frac{{d{\varepsilon _y}}}
{{d{\varepsilon _x}}} =  - \frac{{d{\varepsilon _z}}}
{{d{\varepsilon _x}}}\label{eq:a5}
\end{equation}

Therefore, using the above relations, we have from \cite{poisson_ratio}

%get the following equation

%\begin{equation}
%- \nu_{obj} \int\limits_L^{L - \Delta L} {\frac{{dx}}
%{x}}  = \int\limits_L^{L + \Delta L'} {\frac{{dy}}
%{y} = } \int\limits_L^{L + \Delta L'} {\frac{{dz}}
%{z}}, \label{eq:a6}
%\end{equation}

%which can be simplified as

\begin{equation}
{\left( {1 - \frac{{\Delta L}}
{L}} \right)^{ - \nu_{obj} }} = \left( {1 + \frac{{\Delta L'}}
{L}} \right). \label{eq:a7}
\end{equation}

Combining eqs. (\ref{eq:a7}) and (\ref{eq:a4}), we have

\begin{equation}
\frac{{\Delta A}}
{A} = {\left( {1 - \frac{{\Delta L}}
{L}} \right)^{ - 2\nu_{obj} }} - 1, \label{eq:a8}
\end{equation}

Therefore, due to the applied force and resultant deformation, the new surface area is given by

\begin{equation}
A' = A + \Delta A = A{\left( {1 - \frac{{\Delta L}}
{L}} \right)^{ - 2\nu_{obj} }}, \label{eq:a9}
\end{equation}

where $\Delta L = \frac{F_{surf}}{k_{obj}}$.

\begin{table}[t!]
\small
\vspace{0.6cm}
\caption{Material Selection for Simulations.\label{tbl:sim_mat}}
\begin{center}
\vspace{-0.4cm}
\begin{tabular} {|c|c|c|c|c|c|c|c|c|}
\hline
 & Example & Young's  & Density & \multicolumn{4}{|c|}{Friction Coefficients \cite{fric1,fric2,fric3,fric4,grigoriev1997handbook,jewett2008physics,asm1992friction,blau2008friction,blau2002appendix,hasgalldatabase,farvid2005association,mendez1960density,derler2012tribology,skinthesis}} & \\
%\cline{5-8}
 & of a & Modulus & ($\rho_{obj}$) & \multicolumn{4}{|c|}{} & Ratio\\
\cline{5-8}
General & common & ($E$) & \cite{ashby2008ces,hasgalldatabase} & \multicolumn{2}{|c|}{Wood Surface} & \multicolumn{2}{|c|}{Glass Surface} & ($\nu$)*\\
\cline{5-8}
Material & object & \cite{ashby2008ces,hasgalldatabase} & \cite{density1,density2,density3,witkiewicz2006properties} & Static & Kinetic & Static & Kinetic & \cite{ces}\\ 
\cline{5-8}
 & type & \cite{iivarinen2011experimental,kot2012elastic} & kg/m\textsuperscript{3} & ($\mu_s$) & ($\mu_k$) & ($\mu_s$) & ($\mu_k$) & \cite{birch2009biomechanical}\\
 &  & N/m\textsuperscript{2}  &  &  &  &  &  & \cite{todd1994three}\\ 
\hline
\hline
ABS & Beverage &  1.60E+09 & 1110.00 & 0.5 & 0.4 & 0.3 & 0.25 & 0.41\\
Plastic & containers &  &  &  &  &  &  & \\
\hline
Glass & Beverage & 6.25E+10 & 2250.00 & 0.17 & 0.14 & 0.95 & 0.35 & 0.2\\
 & containers & &  &  &  &  & & \\
\hline
Pine & Furniture & 1.68E+10 & 520.00 & 0.5 & 0.364 & 0.36 & 0.14 & 0.38\\
Wood  &  & &  &  &  &  & & \\
\hline
Ceramic  & Counter &  7.00E+10 & 2250.00 & 0.5 & 0.4 & 0.42 & 0.35 & 0.18\\
  & Tops &  &  &  &  &  & & \\
\hline
Steel  & Appliances & 2.00E+11 & 7850.00 & 0.61 & 0.35 & 0.13 & 0.12 & 0.27\\
 &  &  &  &  &  &  &  & \\
\hline
Polymer & Mattresses & 2.50E+05 & 54.00 & 0.12 & 0.11 & 0.37 & 0.3 & 0.27\\
Foam  & & &  &  &  &  &  & \\
\hline
Light & Pillows & 3.30E+04 & 16.00 & 0.12 & 0.11 & 0.37 & 0.3 & 0.28 \\
Foam  & & &  &  &  &  &  & \\
\hline
Soft & Seat & 1.00E+04 & 48.06 & 0.12 & 0.11 & 0.55 & 0.5 & 0.30 \\
 Foam & Cushions & &  &  &  &  &  & \\
\hline
Open Cell & Sponges & 2.20E+04 & 72.08 & 0.12 & 0.11 & 0.55 & 0.5 & 0.30\\
 Foam &  &  &  &  &  &  &  & \\
\hline
Natural & Footwear & 2.00E+06 & 925.00 & 0.9 & 0.7 & 0.87 & 0.7 & 0.50\\
Rubber &  &  &  &  &  &  &  & \\
\hline
Neoprene & Clothing & 1.35E+06 & 1240.00 & 0.9 & 0.7 & 0.87 & 0.7 & 0.49\\
Rubber & and Bags &  &  &  &  &  &  & \\
\hline
Fat & Human & 1.90E+03 & 919.60 & 0.91 & 0.6 & 0.45 & 0.3 & 0.41\\
(Tissues) & Body &  &  &  &  &  &  & \\
\hline
Muscle & Human & 1.28E+04 & 1060.00 & 0.91 & 0.6 & 0.45 & 0.3 & 0.30\\
(Thigh) & Body & &  &  &  &  &  & \\
\hline
\end{tabular}
\end{center}
\vspace{-0.2cm}
*Ratio = Poisson's Ratio
\end{table}

\section{Experiments with Physics-Based Simulations}\label{sec:sims}
Using our physics-based model from Section \ref{sec:simulations}, we generated simulated data and used the data for our experiments. The simulations with the physics-based model enabled us to generate data with a wide variety of robot, object, and environment conditions, which would be challenging with experiments with a real robot (Section \ref{sec:exps}). We focused on whether our algorithms could classify 
%\ck{it's not clear this is object categorization} 
objects into four different categories and whether the classification performance could generalize to robots having varying mechanical characteristics. By \textit{varying mechanical characteristics}, we mean motion of the robot in which the robot arm's joint stiffness and velocity are varied. 

\subsection{Experimental Setup}\label{ssec:sim_setup}
We modeled the objects as cubes of 10 different volumes. We varied the volumes of the cubes linearly. The edge-length varied from $l_{obj} = 0.01$ m to $l_{obj} = 0.2$ m. We chose 13 different materials to model the objects for our simulations. We chose materials representative of everyday household objects and the human body (See Table \ref{tbl:sim_mat}). We calculated the mass of the object, $m_{obj}$, using eq. (\ref{eq:8}), where $\rho_{obj}$ is the density of the object. To calculate the friction coefficients, we modeled the objects resting on two kinds of surfaces made of wood and glass found on commonly used table tops (See Table \ref{tbl:sim_mat}).

We labeled objects as hard if the calculated stiffness ($k_{obj}$) was greater than $100,000$ N/m which is the stiffness of a medium-stiff environment \cite{daneshmend1990adaptation}. Note, stiffness is a property of object structure as well as its material and is distinct from the Young's modulus of a material which is a material property. For example, a flat thin rectangular block of plastic and a long slender cylindrical plastic bottle may have same Young's modulus (both are made of plastic, assuming the material is homogeneous) but they may have completely different stiffness due to different structural properties. If the calculated stiffness of the object ($k_{obj}$) was less than or equal to $100,000$ N/m, we labeled it as `Soft' (See Section \ref{ssec:limitations} for more detailed discussion on the `hardness' of an object). We labeled objects as `Moved' or `Unmoved' based on whether $x_{obj}$ was greater than $0$ or equal to $0$, respectively, at the end of the simulation. We ran simulations with the physics-based model for $5$s. The mass of the robot, $m_{arm}$, was $1.167$ kg based on the model of the forearm of the robot `Cody' used in our experiments. During the simulations, the robot-arm model came in contact with objects in the shape of a cube of various edge-lengths. Each object of mass $m_{obj}$ is made up of a single material from Table \ref{tbl:sim_mat}.

\begin{equation}
{m_{obj}} = \rho_{obj}{l_{obj}}^3 \label{eq:8}
\end{equation}

For each object, we calculated the stiffness as

\begin{equation}
k_{obj} = 2{l_{obj}}{E_{obj}}, \label{eq:8a}
\end{equation}

where $E_{obj}$ is the Young's modulus of the material of the object. We derived \ref{eq:8a} for an object under both cantilever and axial loading.

\subsection{Experimental Procedure}\label{ssec:sim_proc}
For the experiments, we simulated the robot moving using varying stiffness and velocity. Specifically, we trained the algorithms with three of the four possible combinations of stiffness and velocity conditions (low-velocity-low-stiffness, low-velocity-high-stiffness, high-velocity-low-stiffness, and high-velocity-high-stiffness) and tested with the other combination to find out how well the results generalized to different robot conditions. We repeated this procedure for each of the four conditions and reported the mean classification accuracy for all the conditions. We set the robot's stiffness ($k_{act}$) to a low value ($543$ N/m) or a high value ($2050$ N/m) based on the stiffness values identified in \cite{chen2015evaluation} and the velocity ($\dot{x}_{arm}$) to a low value ($0.005$ m/s) or a high value ($0.02$ m/s). These values correspond with the values used in our experiments with a real robot 
%\ck{with a real robot?} 
in Section \ref{sec:variable_exp}. We set the damping ($b_{act}$) based on a critically damped system as given in eq. (\ref{eq:21a}). 

\begin{equation}
b_{act} = 2\eta\sqrt{{k_{act}m_{arm}}}, \label{eq:21a}
\end{equation}
where $\eta = 1$ is the damping ratio of a critically-damped system \cite{rao2007vibration}.

\subsubsection{Data Collection}\label{sssec:sim_proc_data}
The simulated robot made contact with a set of solid cube objects (See Section \ref{ssec:sim_setup}) made of materials given in Table \ref{tbl:sim_mat} while performing a simple, goal-directed reaching motion as shown in Section \ref{ssec:bot_obj_model}. We simulated the robot-object interactions on both a wooden and a glass table. We simulated the robot movement using an equilibrium-point control similar to that of our experiments with the real robot. We actuated the simulated robot by commanding a goal location and the robot moved according to a PD controller. The final goal location was inside the object. 
%Our objective was to see if the classification performance of our algorithm could generalize across varying motion conditions. We trained the classifiers using the simulated data from a set of stiffness and velocity settings and tested those with a different set of stiffness and velocity settings. 

%\ck{this should be earlier and it should be justified. isn't this the main point of the paper?}
%\ck{but you only report one number. what is the number?} \ck{the whole point of this paper is supposed to be generalization, but the details of the generalization you tested are missing or obscured!} 

Our algorithm classified the objects into the four categories `Hard-Unmoved', `Hard-Moved', `Soft-Unmoved', and `Soft-Moved'. There were 1835 simulation trials with 544 `Hard-Unmoved', 365 `Hard-Moved', 496 `Soft-Unmoved' and 430 `Soft-Moved' trials. Note that for each of the trials in which an object moved, we fixed it to generate one additional `Unmoved' trial, similar to our experiments with the real robot.

\subsubsection{Preprocessing and Feature Selection}\label{sssec:sim_proc_features}
We truncated the data to begin at the estimated onset of contact (whenever the force exceeds $0~N$) between the robot and the object. We collected three time-varying features at each time-instant for $5~s$ at 100 Hz. Our `force' feature is $F_{surf}$ (See Section \ref{ssec:bot_obj_model}), `contact area' feature is $A'$ (See Section \ref{ssec:area_model}), and `motion' feature is $x_{arm}$ (See Section \ref{ssec:bot_obj_model}). Note, if the robot loses contact with the object at any time-instant ($F_{surf} = 0$), $A'$ becomes $A$. However, to match the scenario of experiments with the real robot (See Section \ref{subsubsec:process}), we make the `contact area' feature $0$ for those time-instants during preprocessing. We expected these three features (force, contact area, and motion) to be informative about the object's softness and mobility. These features are similar to the features selected in our experiments with a real-robot (See Section \ref{subsubsec:process}). 

%\subsection{Sensor Model}\label{ssec:sensor_model}
The frequency of the signal from tactile sensing forearm on the real robot is 100 Hz. We modeled an analog-to-digital anti-aliasing filter for the tactile sensing forearm using a low-pass Butterworth filter of order 6 and cut-off frequency of 200 Hz. We also modeled the tactile sensing forearm noise as Gaussian with zero mean and a signal-to-noise ratio of $0.5\%$ for force measurements and zero mean and a signal-to-noise ratio of $0.05\%$ for contact area measurements. We modeled the joint encoder noise as Gaussian with zero mean and a signal-to-noise ratio of $0.05\%$. The signal-to-noise ratio ($SNR$) values correspond with the sensor resolution values on the real robot and we used eq. \ref{eq:21abc} to compute them

\begin{equation}
SNR = \frac{\sigma_{signal}^{2}}{\sigma_{noise}^{2}}. \label{eq:21abc}
\end{equation}

\subsection{Implementation}\label{ssec:implement_sim}
We performed the simulations in MATLAB/Simulink with the `ode15s (stiff/NDF)' solver \cite{solvers} of maximum order `5' using the `Variable-step' solver type and the `Full perturbation' Jacobian method as well as the `Adaptive' zero-crossing option. The resultant simulation trials are of variable length due to the `Variable-step' solver type. To make each simulation trial a vector of uniform length, first, we interpolated the data to a very high sampling rate of 100,000 Hz. Then, we applied a low-pass Butterworth filter to the signal. Finally, we resampled the data to 100 Hz to match the frequency of the tactile sensing forearm, and added Gaussian noise to match sensor noise. 

We implemented continuous univariate and multivariate HMMs as well as LSTMs to model the temporal trends of features for different categories of objects. We modeled each of these four object categories: \textit{Hard-Unmoved}, \textit{Hard-Moved}, \textit{Soft-Unmoved}, and \textit{Soft-Moved}, using an HMM for each category ($\lambda_{HU},\lambda_{HM},\lambda_{SU},\lambda_{SM})$. 

We used the GHMM toolkit \cite{ghmm} to model the HMMs and implemented them in Python. We trained the models with the standard Baum-Welch algorithm, which uses expectation maximization. For testing, we ran the Viterbi algorithm which estimates the maximum likelihood with which a model can describe a given test data. These are standard methods for modeling sequential data (see \cite{Rabiner1990} for details).  We ran the Viterbi algorithm on the given test data for each of the trained HMM models and assigned the category associated with the model that returned the highest likelihood. 

For LSTMs, we used the 'Keras' library \cite{chollet2015keras} with the 'Theano' backend \cite{2016arXiv160502688short}. We trained our model for 10 epochs and used a batch-size of 5.

For comparison purposes, we implemented a one nearest neighbor classifier (1-NN) using the scikit-learn package \cite{scikit-learn} in Python. We used PCA to reduce the dimensionality of the concatenated features similar to \cite{BhattacharjeeRehgKemp2012}. We used the Amazon EC2 cloud computing service \cite{awsec2} to run the experiments on a c3.4xlarge system (high performance compute-optimized instances) with 30 GB of memory, 16 vCPUs and multiple c4.8xlarge systems with 60 GB of memory, 32 vCPUs. All of the systems were 64-bit Ubuntu 14.04 Linux platforms. 

\subsection{Algorithm Parameters}\label{ssec:sim_preprocess}
For the univariate and multivariate HMMs, we analyzed the performance with 10 states. We set a uniform prior to all the states and initialized the emission matrix with Gaussian distributions with means and standard deviations obtained from the training data. For multivariate HMMs using multiple features (force, area, and/or motion), we used a spherical covariance matrix for initialization. Also, for multivariate HMMs, we scaled each feature ($f$) to a scaled feature ($S_f$) according to eq. (\ref{eq:scale}) to normalize the values. 

\begin{equation}
{S_f} = \left( {f - \text{mean}(f)} \right)/\text{std}(f)\; \label{eq:scale}
\end{equation}

For LSTMs, we initialized the parameters with a uniform distribution, used 'softsign' activation functions for the hidden layers, and 'softmax' activation function \cite{chollet2015keras} for the fully connected output layer. Our dropout probability was 0.2. We used 'RMSprop' \cite{chollet2015keras} as the optimizer and 'categorical\_crossentropy' \cite{chollet2015keras} as our loss function because our task is a classification task. We used the 'MinMaxScaler' function \cite{chollet2015keras} to scale multivariate features for LSTMs.

To compare this with our previous work \cite{BhattacharjeeRehgKemp2012}, we used 1-NN. Before using 1-NN, we applied principal component analysis (PCA) representing more than 95\% variance in the training data to reduce the dimensionality as described in our previous work \cite{BhattacharjeeRehgKemp2012}. We used PCA to reduce the effect of noise. As with the HMMs, we used eq. (\ref{eq:scale}) to scale the multivariate features for 1-NNs. 

\subsection{Results}\label{sec:results_sim}
As seen in Table \ref{tbl:summary_table_sim}, multivariate HMMs with contact force and motion as the features performed the best (82.13\%) when compared to our previous method, 1-NN \cite{BhattacharjeeRehgKemp2012}, which failed to generalize (best performance was 64.58\% with area and motion features) across different robot-arm stiffnesses and velocities. Univariate HMMs also failed to generalize. LSTMs showed the best performance with force, area, and motion features (80.46\%).

\begin{figure}[t!]
\centering
\includegraphics[height=7cm]{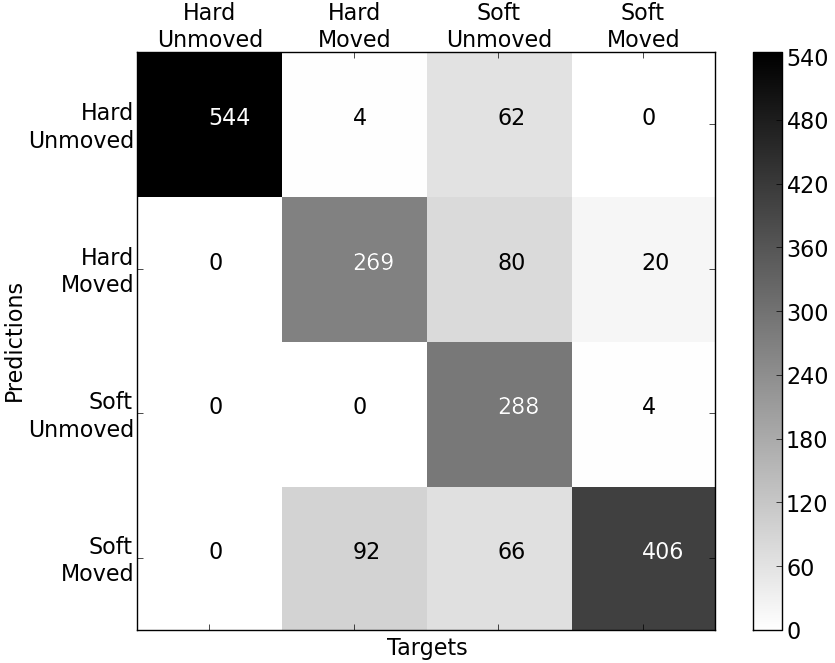}
\caption{\label{fig:hmm_sim_conf}\textit{Classification into four categories using multivariate HMMs with 10 states for experiments with physics-based models. The figure shows the result with force and motion features from the robot moving with varying stiffness and velocity. The numbers in the figure represent the number of trials.}}
\end{figure}

Note that, multivariate features with HMMs and LSTMs showed better performance compared to univariate features. Fig. \ref{fig:hmm_sim_conf} shows the confusion matrix from multivariate HMMs with force and motion features. The algorithm classified 'Hard-Unmoved' category well. But, there is some confusion between 'Soft-Moved', 'Soft-Unmoved' and 'Hard-Moved' categories. Section \ref{ssec:limitations} summarizes a possible reason for this.

\begin{table}[ht!]
\small
\centering
\begin{tabular}[b]{|c|c|c|}
\hline
Algorithm & Features & Ranked Accuracy (\%)\\
\hline
\hline
HMM & f + m & \bf{82.13} \\
\hline
LSTM & f + m + a & 78.02\\
\hline
LSTM & m + a & 70.64 \\
\hline
LSTM & m & 68.64 \\
\hline
LSTM & f + m & 66.73 \\
\hline
LSTM & f + a & 64.68 \\
\hline
1-NN & m & 64.41 \\
\hline
1-NN & m + a & 62.72 \\
\hline
HMM & f + m + a & 59.34\\
\hline
LSTM & f & 58.52\\
\hline
HMM & m & 57.87 \\
\hline
1-NN & a & 57.6 \\
\hline
HMM & m + a & 48.01 \\
\hline
1-NN & f + m & 40.93 \\
\hline
1-NN & f & 40.71 \\
\hline
1-NN & f + m + a & 40.54 \\
\hline
1-NN & f + a & 40.44 \\
\hline
HMM & a & 39.4 \\
\hline
HMM & f & 39.07 \\
\hline
HMM & f + a & 36.89 \\
\hline
LSTM & a & 36.17 \\
\hline
 Majority Classifier & & 29.65\\
\hline
 Random Guess & & 25.0\\
\hline
\end{tabular}
\caption{\label{tbl:summary_table_sim}\textit{Summary of Algorithm Performance for Simulations (Ranked based on Performance). Note `f' = force, `a' = contact area, and `m' = motion feature.}}
\end{table}

\section{Experiments with a Real Robot}\label{sec:exps}
We also performed a set of experiments with a real robot. Similar to our experiments with the physics-based models in Section \ref{sec:sims}, our experiments with the real robot varied arm stiffness and velocity.

\subsection{Experimental Setup}\label{subsubsec:cody}
We used the robot, Cody, for our experiments. Cody, as shown in Fig. \ref{fig:cody}, is a statically stable (wheeled) mobile humanoid robot weighing approximately 160 kg. The components of the humanoid robot are: two Meka A1 arms, a Segway omni-directional base and a Festo 1-DOF (degree of freedom) linear actuator for a spine to adjust the torso height. The two seven-DOF anthropomorphic arms contain series elastic actuators. When we control these arms, each joint simulates a low-stiffness, visco-elastic, torsion spring. We control the robot's arms by changing the equilibrium angles of these simulated springs.

Cody has a force-sensitive skin covering its forearm. Meka Robotics and the Georgia Tech Healthcare Robotics Lab developed the forearm tactile skin sensor, which is based on Stanford's capacitive sensing technology, as described by Ulmen \textit{et al.} \cite{UlmenEdsingerCutkosky2012}. The skin consists of a capacitive pressure-sensor array. We refer to the elements of this array as taxels (tactile pixels). There are 384 taxels on the entire skin, and these are distributed in a 24 x 16 array, with each taxel being 9 mm x 9 mm in size. The array of taxels reports the measured force applied to each taxel at 100 Hz.

\subsection{Experimental Procedure}\label{sec:variable_exp}
We conducted experiments to determine whether the performance of our algorithm could perform well on data collected with different robot arm stiffness and velocity than the training data. For the experiments, we selected two velocity settings, low = $5$ deg/s and high = $20$ deg/s, and two arm stiffness settings, low = $2.01$ Nm/rad and high = $20.1$ Nm/rad.

\subsubsection{Data Collection}\label{subsubsec:data}
\begin{figure*}[t!]
\centering
\begin{tabular}{c}
\includegraphics[height=3.1cm]{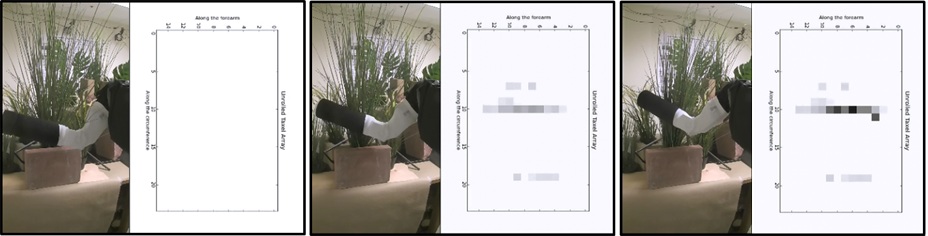}\\
\end{tabular}
\caption{\label{fig:sequence}\textit{Sequence of images that illustrates the data collection for our experiments on inferring mechanical properties of objects (foliage). Each image shows a picture of the robot Cody and a visualization of the data from the forearm skin sensor as a 24 x 16 image (darker pixels correspond to larger forces). The leftmost picture shows a non-contact situation, the middle picture corresponds to the situation just after the onset of contact, and the rightmost picture shows the situation when the robot has pushed the foliage.}}
\end{figure*}

The robot made contact with a set of objects on a wooden table while performing a simple, goal-directed reaching motion. We actuated the robot's shoulder joint only and it pushed into soft and hard objects in moved and unmoved conditions with varied arm stiffness (compliance) and velocities. We performed experiments with the eight objects shown in Fig. \ref{fig:objects_var} (seven in both moved and unmoved conditions, one [heavy iron bucket] in the unmoved condition only). We selected large objects that have mostly uniform material properties and vary widely in their mass, friction, and stiffness. We actuated the robot's shoulder joint by commanding a goal angle in the joint space. The robot arm tried to reach the goal using a joint PD controller. We selected the final goal angle in joint space such that the equivalent point in the Cartesian space was inside the object. Thus, the robot would come in contact with the object before reaching the goal. When the robot incidentally came in contact with the object, it pushed against it and tried to reach the goal as shown in Fig. \ref{fig:sequence}. For each object, we collected haptic data by commanding the same goal angle for the arm and recording the sensor readings from the taxels of the forearm skin at approximately 100 Hz. 

We labeled each of these objects as either soft or hard. For objects that could be pushed aside by the robot's motion, we fixed them with a clamp or a heavy weight so that we could obtain both moved and unmoved conditions. We repeated the experiments for four trials with each of the stiffness and velocity settings. We collected data for a total of 240 trials (($7$ objects x $2$ stiffness x $2$ velocities x $2$ conditions x $4$ trials) + ($1$ object x $2$ stiffness x $2$ velocities x $1$ condition x $4$ trials)). Each object category had 60 trials. Our experiments with the heavy iron bucket were only with the unmoved condition because it could not be moved by the robot's motion.

\subsubsection{Preprocessing and Feature Selection}\label{subsubsec:process}
After recording the time-series data using the forearm taxel array, we truncated them to begin at the estimated onset of contact between the robot and object. We estimated the onset of contact whenever the force exceeded a threshold. We represented the data at every time step as a gray-scale image, as shown in Fig. \ref{fig:sequence}. We converted this image to a binary image, representing the taxels in contact by applying a threshold to each taxel. We used two thresholds. One threshold ($0.01$ N) was for objects which were less likely to harm the robot-arm. For the other set of objects, we used a larger threshold ($0.1$ N) to account for the extra covering that we put over the skin sensor to prevent damage to the skin. This is equivalent to biasing the skin sensor. Then, we computed the connected components on the binary image to segment the contact regions. For the connected component with the largest area, we computed three features.

\begin{figure}[t!]
\centering
\begin{tabular}{c}
\includegraphics[height=4.5cm]{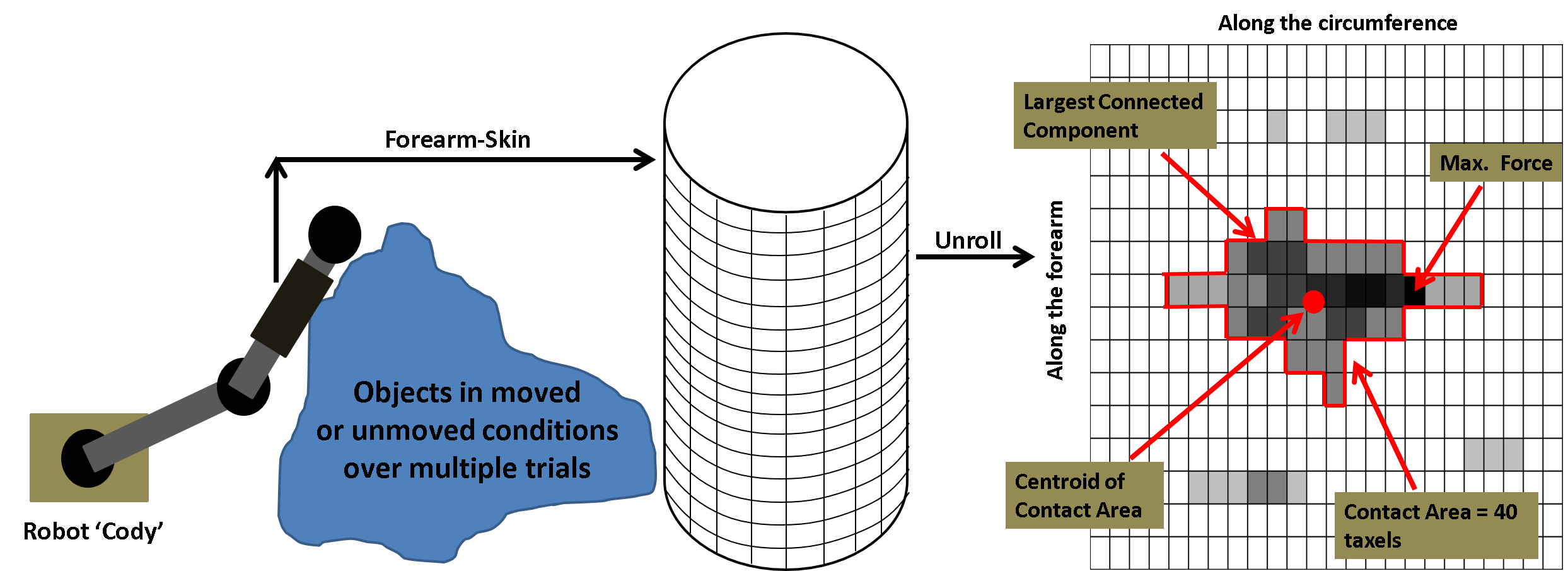}
\end{tabular}
\caption{\label{fig:protocol}\textit{Schematic representation of the experimental protocol.}}
\end{figure}

Figure \ref{fig:protocol} depicts the feature collection procedure. The first feature is the maximum force ($F_{max}$) measured by a taxel in the contact region at each time step. This is analogous to measuring the highest pressure. In our Initial tests, the maximum force performed better than total force ($F_{total}$) or mean force ($F_{mean}$). 
%This is probably because total force is unable to clearly differentiate between cases of small contact forces over a large area with cases of large contact forces over a small area. Mean force in a contact area is also unable to distinguish large forces if the surrounding area has low forces. 

For the second feature, we estimated the area of contact ($a$) between the arm and object (contact region) as the number of taxels in the connected component. 

For the third feature, we estimated the distance that the centroid of the connected component traveled in the world frame from its position at the onset of contact ($d$). We held the robot's torso fixed throughout the trials and used the forward kinematics from the robot's torso to the center of the contact location on the robot's forearm to estimate the 3D positions and distance traveled. 

Similar to the experiments with physics-based models, we expected these three features to be informative about the object's softness and mobility. For example, with increasing force applied to a soft, unmoved object, we would expect the contact area to increase. Likewise, we would expect the 3D position of the contact area to travel when encountering moved and soft objects. When making contact with a hard and unmoved object, we would expect the maximum force to increase. Our algorithms used the values of maximum force, the number of taxels in the contact region, and the contact motion for each trial during the first $1.2$s time window after the onset of contact.

\subsection{Implementation}\label{ssec:exp_implement}
We implemented HMMs, LSTMs, and 1-NN similar to our experiments with physics-based models in Section \ref{ssec:implement_sim}. We performed experiments with the real robot using a 32-bit Ubuntu 10.04 system with 8 CPU(s) and Intel(R) Core(TM) i7-2600 CPU processor with 3.40GHz. We used Python to send commands to the robot's real-time PC using ROS-Diamondback~\cite{ros}.

\subsection{Algorithm Parameters}\label{ssec:exp_params}
We used 10 states for the univariate and multivariate HMMs. We set a uniform prior for all states and initialized the emission matrix with Gaussian distributions with means and standard deviations obtained from the data.

For LSTMs, similar to experiments with physics-based models, we initialized the parameters with a uniform distribution, and used the same structure, scaling function, activation functions and dropout probability.

To compare this with our previous work~\cite{BhattacharjeeRehgKemp2012}, we extracted the features and converted them to a low-dimensional representation of these feature vectors using PCA. We used a dimensionality of three which represented greater than 95\% of the variance in the data. 
%Note that the labels in our previous work~\cite{BhattacharjeeRehgKemp2012} were `Rigid-Fixed', `Rigid-Movable', `Soft-Fixed', and `Soft-Movable' which are different from the labels in our current work which are `Hard-Unmoved', `Hard-Moved', `Soft-Unmoved', and `Soft-Moved'.

Also, we used the same evaluation procedure as in the experiments with the physics-based model (Section \ref{ssec:sim_preprocess}). We trained on three conditions of robot stiffness and velocity and tested on one. We repeated this procedure for each of the four conditions. For multivariate HMMs, we scaled each feature ($f$) to a scaled feature ($S_f$) according to eq. (\ref{eq:scale}) to normalize the values.

\begin{figure}[t!]
\centering
\includegraphics[height=4.7cm]{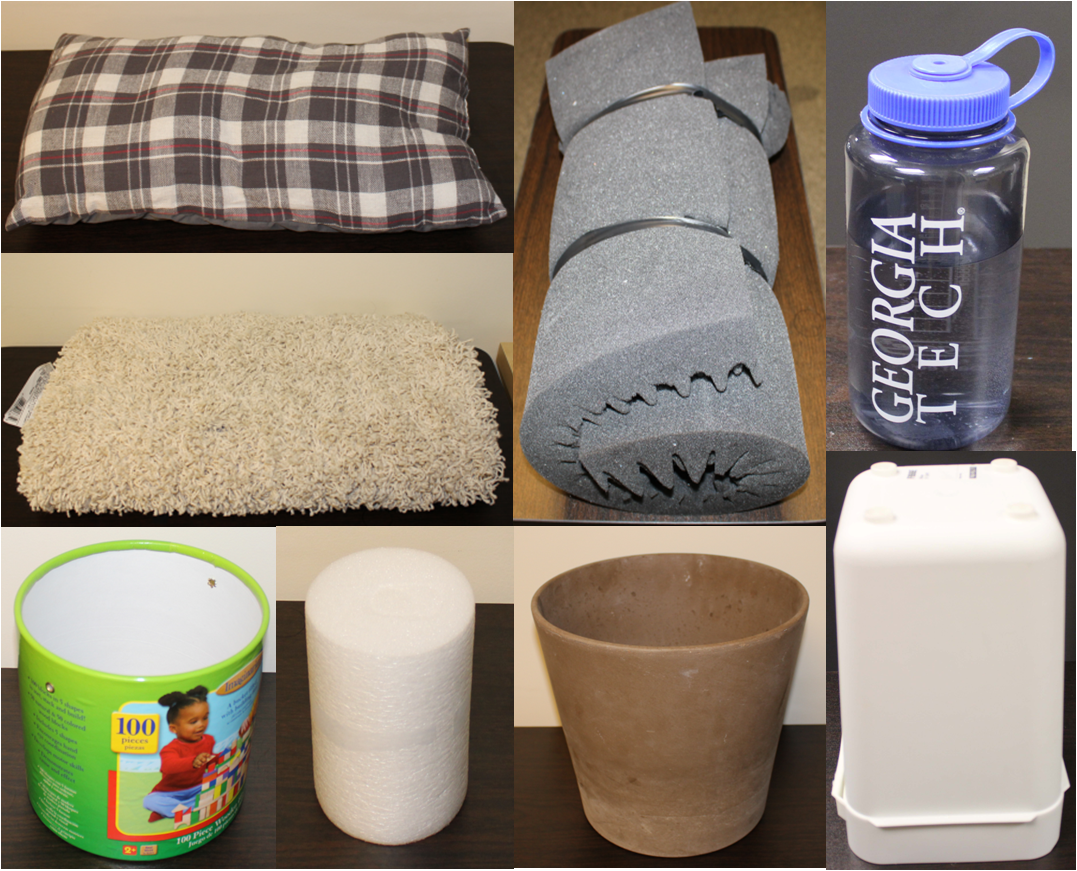}
\includegraphics[height=5cm]{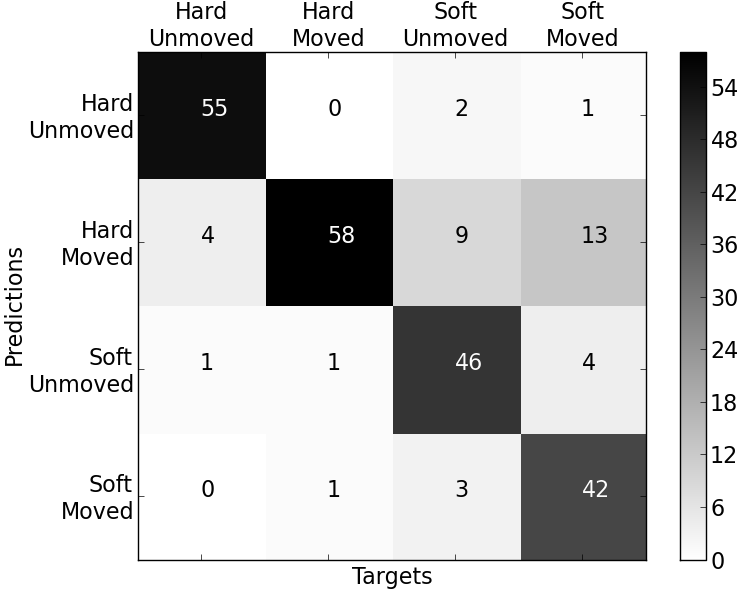}
\caption{\label{fig:objects_var}\textit{Set of objects for experiments with varying stiffness and velocity (Left). Classification into four categories using multivariate HMMs for experiments using the robot `Cody'. The figure shows the results with force, area, and motion features from the robot moving with varying stiffness and velocity. The numbers in the figure represent the number of trials (Right).}}
\end{figure}

%%%%%%%%%%%%%%%%%%%%%%%%%%%%%%%%%%%%%%%%%%%%%%%%%%%%%%%%%%%%%%%%%%%%%%%%%%%%%
%%
%%  SECTION : Results and Discussion
%%
%%%%%%%%%%%%%%%%%%%%%%%%%%%%%%%%%%%%%%%%%%%%%%%%%%%%%%%%%%%%%%%%%%%%%%%%%%%%%
\subsection{Results}\label{sec:results}
We used both univariate HMMs, multivariate HMMs, and LSTMs for classification to model the temporal trends of all combinations of the three feature vectors: maximum force ($F_{max}$), contact area ($a$) and contact motion ($d$). Table \ref{tbl:summary_table_exp} presents the results for classification into four categories: 1) \textit{Hard-Unmoved}, 2) \textit{Hard-Moved}, 3) \textit{Soft-Unmoved}, and 4) \textit{Soft-Moved}. Our previous method (from \cite{BhattacharjeeRehgKemp2012}) performed poorly, and the highest accuracy was only 37.92\% with a single feature (motion) and only 35\% with three features (force, area, and motion) using a dimensionality of three. Note that 12 trials could not be captured up until the time window of 1.2 s because of the varying velocity conditions. In those cases, we extrapolated the data with the mean value for that particular trial to obtain a consistent time window of 1.2 s. With HMMs using a single feature (force or area), the accuracy improved only slightly to 40.41\%. However, using multivariate HMMs, the accuracy improved to 83.75\% with all three features. This provides evidence that multivariate HMMs can be used to generalize the data-driven inference results to testing data that differ from training data due to varying robot conditions. Fig. \ref{fig:objects_var} (Right) shows the resulting confusion matrix. From the figure, we see that some `Soft-Moved' and `Soft-Unmoved' trials were categorized as `Hard-Moved'. However, the algorithm categorized `Hard-Moved' and `Hard-Unmoved' categories well. Note, the results with LSTMs did not perform well. With access to more data, LSTMs might match the better results we obtained with physics-based models shown in Section \ref{sec:results_sim}. 

% \begin{table}[t!]
% \small
% \vspace{0.6cm}
% \caption{Summary of Algorithm Performance for Experiments.\label{tbl:summary_table_exp}}
% \begin{center}
% \vspace{-0.4cm}
% \begin{tabular} {|c|c|c|}
% \hline
%  Algorithm & Features & Accuracy\\
%  &  & (\%) \\
% \hline
% \hline
%  & Force & 27.5\\
% \cline{2-3}
%  & Area & 26.25\\
% \cline{2-3}
%  Previous & Motion & 37.92\\
% \cline{2-3}
%  Method (k-NNs) \cite{BhattacharjeeRehgKemp2012} & Force, Area & 39.17\\
% \cline{2-3}
%  & Force, Motion & 25.42\\
% \cline{2-3}
%  & Area, Motion & 28.33\\
% \cline{2-3}
%  & Force, Area, Motion & 35.0\\
% \cline{1-3}
%  & Force & 40.41 \\
% \cline{2-3}
%  Univariate & Area & 40.41\\
% \cline{2-3}
%  HMMs & Motion & 32.5 \\
% \cline{1-3}
%  & Force, Area & 52.08 \\
% \cline{2-3}
%  Multivariate & Force, Motion & 71.25 \\
% \cline{2-3}
%  HMMs & Area, Motion & 60.41 \\
% \cline{2-3}
%  & Force, Area, Motion & \bf{83.75}\\
% \cline{1-3}
%  Random Guess&  \multicolumn{2}{|c|}{25.0} \\
% \cline{1-3}
%  Majority Classifier &  \multicolumn{2}{|c|}{25.0} \\
% \hline
% \end{tabular}
% \end{center}
% \vspace{-0.2cm}
% \end{table}

\begin{table}[ht!]
\small
\centering
\begin{tabular}[b]{|c|c|c|}
\hline
Algorithm & Features & Ranked Accuracy (\%)\\
\hline
\hline
HMM & f + m + a & \bf{83.75} \\
\hline
HMM & f + m & 71.25 \\
\hline
HMM & m + a & 60.41 \\
\hline
HMM & f + a & 52.08 \\
\hline
LSTM & f + m & 43.33 \\
\hline
LSTM & f & 40.42 \\
\hline
HMM & f & 40.41 \\
\hline
HMM & a & 40.41 \\
\hline
LSTM & f + a & 40.0 \\
\hline
1-NN & f + a & 39.17 \\
\hline
LSTM & a & 39.17 \\
\hline
1-NN & m & 37.92 \\
\hline
LSTM & f + m + a & 35.83 \\
\hline
1-NN & f + m + a & 35.0 \\
\hline
LSTM & m + a & 34.58 \\
\hline
HMM & m & 32.5 \\
\hline
LSTM & m & 31.67 \\
\hline
1-NN & m + a & 28.33 \\
\hline
1-NN & f & 27.5 \\
\hline
1-NN & a & 26.25 \\
\hline
1-NN & f + m & 25.42 \\
\hline
Majority Classifier & & 25.0\\
\hline
Random Guess & &  25.0\\
\hline
\end{tabular}
\caption{\label{tbl:summary_table_exp}\textit{Summary of Algorithm Performance for Experiments (Ranked based on Performance). Note `f' = force, `a' = contact area, and `m' = motion feature.}}
\end{table}

\section{Discussion and Limitations}\label{ssec:limitations}
Our data and models will be available publicly as a part of our `Open-Access-Haptic-Database (OAHD)'. From the overall results, we learned that:

\begin{itemize}
	%\item{Object-centric state-based methods such as multivariate HMMs performed better than k-NN that explicitly model time, which use a similarity measure between two time-series vectors,}
    \item{It is feasible to use data-driven methods to infer object properties from contact during a reaching motion,}
    \item{Classification results using HMMs and LSTMs (with sufficient data availability) with multiple features can generalize well to different robot behaviors, such as robot-arm stiffness and arm velocity used to collect training data.}
\end{itemize}

We used the physics-based model to do experiments with objects with a wide-variety of stiffnesses. Collecting haptic data from robots touching real-world objects is challenging. The physics-based model we used in this work can help collect data from a wide variety of objects and robot settings (varied robot stiffnesses, velocities etc.) by leveraging the widely available  material properties in online databases. In addition to collecting the real-world data, the results from our physics-based model matched quite well with the results from our real-world data. For example, for both the experiments with physics-based models as well as the real-robot, we note that there is confusion between `hard-moved', `soft-unmoved' and `soft-moved' categories. This is probably because, given the features we have, the algorithm finds it difficult to disambiguate between sliding motion in hard objects and motion due to deformation of soft objects. 

Note that for our experiments with the real robot using multivariate HMMs, there was very little confusion between hard and soft objects in the `Unmoved' condition. However, for experiments with the physics-based models, there was some confusion between hard objects and soft objects. This could be because objects in the hard category for the experiments with the real robot were much stiffer than objects in the soft category. We used a compression spring to compress the objects used in the real-robot experiment and found that the soft objects had stiffnesses ranging between $630~N/m$ to $1500~N/m$ whereas the hard objects had stiffnesses ranging between $8000~N/m$ to $100,000~N/m$ and above. This is not the case for experiments with the physics-based models. We experimented with a wide variety of stiffnesses and used a stiffness threshold of $100,000 ~N/m$ that differentiated hard and soft categories. Thus, the stiffnesses of the objects in the hard and soft categories were much closer (e.g. our simulation labeled an object with stiffness $100,100~N/m$ as hard but labeled an object with stiffness $99,900~N/m$ as soft) for experiments with the physics-based models and this led to more confusion.

For our experiments with physics-based models, we used stiffness as a criterion for labeling objects as hard or soft. In material sciences community, `hardness' is defined as a measure of how resistant an object is to deformation when a compressive force is applied \cite{tabor2000hardness, oliver1992improved}. We are interested in forces only in the elastic deformation range of a material and for elastic deformation ranges, this highly correlates with stiffness of an object \cite{tabor2000hardness, oliver1992improved} which is a function of the object material as well as structural properties. This is related to deformation of the object, as stiffness is monotonically related to deformation. Similarly, we labeled the objects for experiments with the real robot depending on how an object deforms in macroscopic scale due to applied forces. The human labeling of hard vs. soft objects for experiments with the real robot was coarse. However, note that when consistent force was applied, all the objects labeled soft showed larger macroscopic deformation when compared to objects labeled hard.

Finally, we intentionally made the problem harder by not giving the perceptual classifiers information about the robot's joint stiffness or arm velocity. This is because robots might not have good stiffness estimates for all contact locations on their bodies. Likewise, robots might have uncharacterized compliant coverings or components like soft robots. Similarly, accurate velocity estimates may not be attainable. Also, although the performance with LSTMs (with sufficient data) and multivariate HMMs generalized well to new robot behaviors used to collect training data, there are some limitations to the results presented here. In this work, our objective was to see if our algorithms can infer haptic properties using simple motions (e.g. linear) without haptic exploratory behaviors. Note that many motions are locally linear and therefore, this simple type of motion may be applicable in many scenarios. However, there are many factors which are relevant to real-world incidental contact like the impact dynamics, non-ideal contact (partial, non-normal) due to different robot trajectories and object shapes and sizes etc. In this paper, we focused on two aspects --- `stiffness' and `velocity'--- which affect the way contact occurs. Other aspects of non-ideal contact merit consideration in the future.

\section{Conclusion}\label{sec:conclusion}
We developed algorithms to infer object properties using haptic information obtained from contact between a robot's tactile sensing forearm and objects in the robot's environment. We showed that using our algorithms and relevant tactile sensing features, haptic inference can be generalized to data collected using different robot-arm velocities and stiffnesses. Our algorithms classified objects into four categories: 1) \textit{Hard-Unmoved}, 2) \textit{Hard-Moved}, 3) \textit{Soft-Unmoved}, and 4) \textit{Soft-Moved}. 

We developed an idealized physics-based model and generated simulated data under varying robot stiffness and velocity. We also performed experiments with a real robot. We used univariate and multivariate HMMs and long short-term memory (LSTM) networks to classify objects under these conditions and compared the results with our previous method of PCA + 1-NN \cite{BhattacharjeeRehgKemp2012}. Our results showed that HMMs are a useful tool to model robot-object interactions. Multivariate HMMs consistently performed better in all cases with varying robot velocity and compliance parameter values and outperformed our previous technique using PCA + 1-NN \cite{BhattacharjeeRehgKemp2012} (See Fig. \ref{tbl:summary_table_sim} and Table \ref{tbl:summary_table_exp}). With the availability of more data, the classification performance using LSTMs can also generalize to data collected using different robot actions. Also, for HMMs and LSTMs, classification results using a combination of relevant features such as force, area, and motion generalize better than using single features. 
%This finding suggests that our methods using multiple features with HMMs and LSTMs (with sufficient data) can continue to perform well under circumstances that differ from training, such as different control methods and arm motions.

\section*{Acknowledgements}
We gratefully acknowledge the support from DARPA's Maximum Mobility and Manipulation (M3) program, Contract W911NF-11-1-603, NSF Emerging Frontiers in Research and Innovation (EFRI) 1137229, and NSF Career Award 1150157. We thank Joshua Wade for his help with selecting the materials for our simulations, as well as Ariel Kapusta and Zackory Erickson for their valuable feedback. We also thank Mark Cutkosky and the Stanford Biomimetics and Dexterous Manipulation Lab for their contributions to the forearm tactile skin sensor.

\vspace*{-7pt}   %ONLY NECESSARY


\bibliographystyle{ws-ijhr}
\bibliography{TRO_Haptic_Classification}


%FIRST AUTHOR
\vspace*{13pt}
\noindent%
\parbox{5truein}{
\begin{minipage}[b]{1truein}
\centerline{{\psfig{file=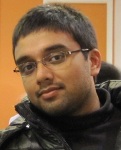,width=1in,height=1.25in}}}
\end{minipage}
\hfill         %to 2nd column
\begin{minipage}[b]{3.85truein}
\small
{{\bf Tapomayukh Bhattacharjee} received the B.Tech. degree from the Department of Mechanical Engineering, National Institute of Technology, Calicut, India, and the M.S. degree from the Department of Mechanical Engineering, Korea Advanced Institute of Science and Technology, Daejeon, Korea. He is currently a Robotics Ph.D. Student
in the Georgia Institute of Technology, Atlanta. He also worked as a Visiting
Scientist with the Interaction and Robotics Research Center, Korea Institute
of Science and Technology, Seoul, Korea. His research interests include
haptic perception, control systems for robotic manipulation, human-robot interaction, machine learning, and teleoperation systems. He is a member of IEEE, IEEE RAS, and IEEE CSS.}
\end{minipage} } %close for parbox

%SECOND AUTHOR
\vspace*{13pt}  
\noindent%
\parbox{5truein}{
\begin{minipage}[b]{1truein}
\centerline{{\psfig{file=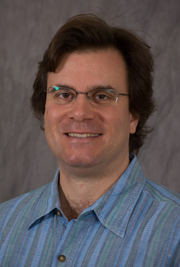,width=1in,height=1.25in}}}
\end{minipage}
\hfill %to 2nd column
\begin{minipage}[b]{3.85truein}
\small
{{\bf James M. Rehg} is a Professor in the School of Interactive Computing at the Georgia Institute of Technology, where he is co-Director of the Computational Perception Lab and is the Associate Director for Research in the Center for Robotics and Intelligent Machines (RIM@GT). He received his Ph.D. from CMU in 1995 and worked at the Cambridge Research Lab of DEC (and then Compaq) from 1995-2001, where he managed the computer vision research group. He received an NSF CAREER award in 2001 and a Raytheon Faculty Fellowship from Georgia Tech in 2005. His research interests include computer vision, medical imaging, robot perception, machine learning, and pattern recognition. Dr. Rehg is currently leading a multi-institution effort to develop the science and technology of Behavior Imaging— the capture and analysis of social and communicative behavior using multi-modal sensing, to support the study and treatment of developmental disorders such as autism.}
\end{minipage} } %close for parbox

%THIRD AUTHOR
\vspace*{13pt}  
\noindent%
\parbox{5truein}{
\begin{minipage}[b]{1truein}
\centerline{{\psfig{file=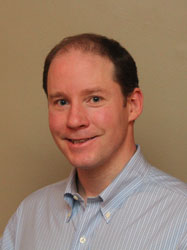,width=1in,height=1.25in}}}
\end{minipage}
\hfill %to 2nd column
\begin{minipage}[b]{3.85truein}
\small
{{\bf Charles C. Kemp} (Charlie) is an Associate Professor at the Georgia Institute of Technology in the Department of Biomedical Engineering with adjunct appointments in the School of Interactive Computing and the School of Electrical and Computer Engineering. He earned a doctorate in Electrical Engineering and Computer Science (2005), an MEng, and BS from MIT. In 2007, he founded the Healthcare Robotics Lab ( http://healthcare-robotics.com ). His lab focuses on mobile robots for intelligent physical assistance in the context of healthcare. He has received a 3M Non-tenured Faculty Award, the Georgia Tech Research Corporation Robotics Award, a Google Faculty Research Award, and an NSF CAREER award. He was a Hesburgh Award Teaching Fellow in 2017. His research has been covered extensively by the popular media, including the New York Times, Technology Review, ABC, and CNN.}
\end{minipage} } %close for parbox

\vfill\eject

\end{document}